\documentclass[acmlarge]{acmart}
\usepackage{multirow}
\usepackage{algorithm}
\usepackage{algpseudocode}

\def \paperNameActsafe {\textit{ActSafe}}
\def \fancyModelName {\textit{HERBERT}}
\AtBeginDocument{%
  \providecommand\BibTeX{{%
    \normalfont B\kern-0.5em{\scshape i\kern-0.25em b}\kern-0.8em\TeX}}}

\setboolean{@printcopyright}{false}
\setboolean{@printpermission}{false}
\setboolean{@acmowned}{false}

\begin{document}

\title{ActSafe: Predicting Violations of Medical Temporal Constraints for Medication Adherence}

\author{Parker Seegmiller}
\email{matthew.p.seegmiller.gr@dartmouth.edu}
\author{Joseph Gatto}
\author{Abdullah Mamun}
\author{Hassan Ghasemzadeh}
\author{Diane Cook}
\author{John Stankovic}
\author{Sarah Masud Preum}
\renewcommand{\shortauthors}{Seegmiller et al.}

\begin{abstract}
Prescription medications often impose temporal constraints on regular health behaviors (RHBs) of patients, e.g., eating before taking medication. Violations of such medical temporal constraints (MTCs) can result in adverse effects. Detecting and predicting such violations before they occur can help alert the patient. We formulate the problem of modeling MTCs and develop a proof-of-concept solution, \paperNameActsafe{}, to predict violations of MTCs well ahead of time. \paperNameActsafe{} utilizes a context-free grammar based approach for extracting and mapping MTCs from patient education materials. It also addresses the challenges of accurately predicting RHBs central to MTCs (e.g., medication intake). Our novel behavior prediction model, \fancyModelName{}, utilizes a basis vectorization of time series that is generalizable across temporal scale and duration of behaviors, explicitly capturing the dependency between temporally collocated behaviors. Based on evaluation using a real-world RHB dataset collected from 28 patients in uncontrolled environments, \fancyModelName{} outperforms baseline models with an average of 51\% reduction in root mean square error. Based on an evaluation involving patients with chronic conditions, \paperNameActsafe{} can predict MTC violations a day ahead of time with an average F1 score of 0.86.

\end{abstract}

\maketitle

\section{Introduction}
\label{intro_section}
\textbf{Medication adherence and regular health behavior: }
According to the CDC, over 48\% of the US population use at least one prescription medicine, and 24\% take three or more \cite{CDCPrescription}.
However, approximately four out of five new prescriptions are filled, and half of them are administered inappropriately \cite{osterberg2005adherence}. Such non-adherence can be attributed to not adhering to the time, dosage, frequency, or duration of medication intake, and potential temporal dependencies with other regular health behaviors (e.g., \textit{eating}, \textit{exercising}, \textit{sleeping}), such as taking medication on an empty stomach or taking medication with food \cite{osterberg2005adherence, ingersoll2008impact, ferguson2017barriers, klakegg2018assisted}. We refer to the temporal constraints associated with medications as \textit{\textbf{M}edical \textbf{T}emporal \textbf{C}onstraints} (\textbf{MTC}s). 

Non-adherence to prescription medications is linked to higher hospital admission rates, increased morbidity, higher healthcare expenses, poor health outcomes, and even death \cite{dimatteo2004variations, chisholm2012cost, CDCPrescription2, ferguson2017barriers, klakegg2018assisted}. In particular, the effect of violating MTCs can range from minor discomfort to emergency room visits \cite{pham2011national}. For example, suppose a patient diagnosed with diabetes and prescribed to take insulin violates critical MTCs regarding insulin. This violation might impact the proper dosing of medication due to not following the proper timing. Subsequently, they can experience hyperglycemia or hypoglycemia and a range of symptoms, including thirst, increased urination, blurred vision, rapid heartbeat, and headache. Similarly, violation of MTCs corresponding to prescription medications for cardiovascular diseases (e.g., ACE inhibitors) can cause various symptoms, including anxiety, chest pain, seizure, or even heart attack \cite{nasution2006use, mayoCACE}.

\textbf{Factors contributing towards violations of medical temporal constraints:}
Adhering to medical temporal constraints is particularly challenging for people chronic diseases and their caregivers because they often have an increased number of medications \cite{ferguson2017barriers,dolce1991medication,donnan2002adherence,paes1997impact, ingersoll2008impact, mayer2001strategies, klakegg2018assisted}. Increased medications leads to poor adherence for several common chronic conditions, including COPD and other respiratory issues \cite{dolce1991medication, ingersoll2008impact}, Diabetes \cite{donnan2002adherence, paes1997impact}, cardiovascular conditions \cite{ferguson2017barriers}, and HIV \cite{ingersoll2008impact, mayer2001strategies}. 
Research suggests that 40\% of non-adherence behavior can be attributed to forgetfulness where an individual forgets to take their medications in the proper way \cite{expressScripts}. 
In addition, these chronic conditions often lead to cognitive impairment. Patients with cognitive impairment and complex dosing schedules showed the poorest rates of adherence in multiple studies \cite{hinkin2002medication, ingersoll2008impact}. Non-adherence to treatment can impact individuals at any age \cite{leven2017medication, burnier2020hypertension}. 


 \textbf{Effectiveness of medication reminder apps to maintain medical temporal constraints}: 
Although medication reminders via mobile applications yield positive results in improving adherence initially, their effectiveness drops over time due to multiple factors including static messaging and failure to address underlying reasons behind low adherence \cite{singh2019reminders}. Typical medication reminders overlook patients' regular health behavior (or activities) and lifestyle factors that impact their medication adherence. Thus they often fail to provide personalized nudges to prevent potential lack of medication adherence resulting from violation of medical temporal constraints (MTCs). Several lifestyle factors impact medication adherence by impacting adherence to the MTCs. For example, when an individual is prescribed to take a medication on an empty stomach, they are required to not eat within about 2 hours of taking the medication. This MTC will impact the daily routine of the individual, possibly leading to lack of medication adherence. In addition, a new prescription can lead to one or more side effects such as change in appetite, sleep, mood, or cognition. Such side effects can contribute to violations of MTCs. Finally, there might be sudden or unexpected changes on one's daily routine that might be overlooked by typical medication reminder apps. For example, waking up late on the weekend can delay the medication intake time and require an individual to reconsider the temporal dependency with other regular health related behaviors (RHBs), e.g., eating.

\textbf{Scope of \paperNameActsafe{} to maintain medical temporal constraints}:
This paper focuses on developing a system that is (i) aware of the medical temporal constraints (MTCs) for regular health behaviors (RHBs) and (ii) capable of accurately predicting potential violations of these MTCs ahead of time. This prediction of potential violations of MTCs can aid the existing cognitive orthotics \cite{ha2014towards, sonntag2015kognit, pollack2002pearl}, behavior reminders \cite{rajanna2014step, dahmen2018smart, emi2017quactive}, or digital personal assistants (e.g., \textit{Alexa, Siri, Google Home}) to alert a patient or their care-providers \cite{klakegg2018assisted} about these violations and subsequently design a resolution scheme. Especially with the surge of popularity of in-home, life-wide personal assistants like Amazon Alexa, such predictive recommendations or system alerts about potential violations of MTCs can be integrated into these personal assistants. Since the ultimate goal of this system is to help patients to \textit{act safely} while following a medical treatment, we call the proposed system \paperNameActsafe{}. This paper addresses the following major challenges.

\begin{itemize}
    
    \item Although \textit{\textbf{M}edical \textbf{T}emporal \textbf{C}onstraints} (MTCs) are critical for medical safety and treatment adherence, to the best of our knowledge, there is no existing solution to formulate and model patient-centric MTCs from drug usage guidelines (also known as patient handouts). This requires detecting the the drug usage guideline that contain one or more MTCs, creating a flexible and robust computational representation of MTCs, and mapping natural language descriptions of MTCs to the computational representation. MTCs vary in terms of temporal precision (i.e., definitive or imprecise) and can involve more than one regular health behaviors (e.g., taking medication \textit{m} hours before having a meal). 
    

    \item  To predict MTC violations, one needs to predict occurrence of relevant regular health behaviors (RHBs) as well. To predict a potential violation of MTCs involving both breakfast and medication intake, we must predict both RHBs, i.e., \textit{breakfast} and \textit{medication intake}. However, existing solutions for activity prediction cannot be directly applied for this task for the following reasons. First, existing activity prediction solutions often either overlook or do not provide reasonably accurate prediction for short-span RHBs like taking medication \cite{minor2015data}. Second, they do not scale well for longer prediction horizon \cite{preum2015maper}, which is crucial for this safety-critical solution to allow room for resolving the potential MTC violations. 
    

    \item Predicting MTC violations well ahead of time can enable us to generate context-aware reminders for medication adherence. However, predicting violations of MTCs requires aligning and fusing the following multimodal health data: (i) time series data of predicted or recognized behavior or activity sequence and (ii) unstructured textual descriptions of MTCs in natural language found in drug usage guidelines. There is no existing solution to align such multimodal health data streams. 
    
    
    
\end{itemize}



The main contributions of this paper are as  follows:
\begin{itemize}
    \item We formulate the problem of medical temporal constraints (MTCs) on regular human behaviors (RHB) for treatment adherence and health safety. We develop a taxonomy of potential MTCs and a context-free grammar (CFG) based model to normalize MTCs from unstructured free-format descriptions. The taxonomy and the CFG model are applied to a real prescription dataset to model MTCs overlapping with RHBs in the context of chronic disease management. 
    We develop and evaluate a novel MTC extraction pipeline to extract MTCs from unstructured free-format drug usage guidelines and map them to our CFG model. This pipeline extracts five types of MTCs that are found in our dataset with F1-scores as high as 0.87. 

    \item Accurately predicting relevant, short-span RHBs with a robust prediction horizon is central to predicting violations of MTCs . We develop the HEalth Related BEhavior pRedicTion model (\fancyModelName{}) that addresses this challenge. The \fancyModelName{} model addresses the varying temporal scale and duration of RHBs through the generalizable process of basis vectorization. The \fancyModelName{} model utilizes a personalized training process over several weeks of high-level RHB labels with behavior start timestamp and behavior duration to predict future RHB timestamps with a robust prediction horizon. Based on evaluation conducted using a real RHB dataset consisting of $28$ patients, \fancyModelName{} yields an average of 51\% reduction in root mean square error (RMSE) over other baselines. We also demonstrate the effectiveness of \fancyModelName{} when predicting RHB across sparse RHB label spaces (i.e., missing data), and irregular patient schedules.
    
    \item We develop a novel rule-based solution to detect and predict violations of different types of MTCs well ahead of time. This solution fuses information from multimodal health data, i.e., predicted behavior time series and unstructured text describing MTCs. This solution is evaluated using a real RHB dataset collected from $28$ chronic disease patients.Each patient is diagnosed with atherosclerotic cardiovascular disease and prescribed to take ACE inhibitors daily. The evaluation demonstrates that \paperNameActsafe{} can predict violations of three MTCs with a weighted average F1 score of 0.86.
    

    
\end{itemize}

\begin{table}
\caption{\textbf{Medical Temporal Constraint (MTC) Examples}: Examples of medical temporal constraints (MTCs) on daily activities imposed by a set of medications or drugs that are commonly used to treat different chronic diseases. The first and second columns show the drug name and the corresponding diseases that they are used to treat. The third column shows drug usage guidelines or advice that are suggested to patients to ensure the drug is administered properly. The fourth column shows the health behaviors/activities mentioned in the advice. All these drugs are selected from the real prescription data used in this paper. The drug usage guidelines are collected from Medscape
\cite{Medscape}.}
\label{table_DUGAdvice}
\begin{tabular}{cclc}
\toprule
Drug                                                            & Relevant Disease                                                                                        & \multicolumn{1}{c}{Drug Usage Guideline}                                                                                                                                                                                                                                                                                                                                                                                                                                 & Relevant RHBs                                                                                           \\ \hline

Desyrel                                                         & \begin{tabular}[c]{@{}c@{}}Depression, Anxiety, \\ Sleep, Pain\end{tabular}                             & \begin{tabular}[c]{@{}l@{}}If drowsiness is a problem and you are \\ taking 1 dose daily, take it at bedtime.\end{tabular}                                                                                                                                                                                                                                                                                                                                               & taking before sleep                                                                                     \\ \midrule
Fentanyl                                                        & Severe pain                                                                                             & \begin{tabular}[c]{@{}l@{}}Avoid activities that might cause your \\ body temperature to rise. (Such as doing \\ strenuous work/exercise in hot weather).\end{tabular}                                                                                                                                                                                                                                                                                                   & \begin{tabular}[c]{@{}c@{}}hard work; \\ hard exercise\end{tabular}                                     \\ \midrule
\begin{tabular}[c]{@{}c@{}}Hydrochl-\\ orothiazide\end{tabular} & \begin{tabular}[c]{@{}c@{}}Hypertension, \\ Congestive \\ heart failure\end{tabular}                    & \begin{tabular}[c]{@{}l@{}}It is best to avoid taking this medication \\ within 4 hours of your bedtime to prevent \\ having to get up to urinate.\end{tabular}                                                                                                                                                                                                                                                                                                          & get up to urinate                                                                                       \\ \midrule
Ambien                                                          & Insomnia                                                                                                & \begin{tabular}[c]{@{}l@{}}Do not take a dose of this drug unless \\ you have time for a full night’s sleep of \\ at least 7 to 8 hours. If you have to wake \\ up before that, you may have some memory \\ loss and may have trouble safely doing \\ any behavior that requires alertness, such \\ as driving. or operating machinery. You \\ may feel alert, but this medication may \\ continue to affect your thinkng making \\ such activities unsafe.\end{tabular} & \begin{tabular}[c]{@{}c@{}}sleeping for \\ 7-8 hours; \\ driving; \\ using machinery\end{tabular}       \\ \bottomrule
\end{tabular}
\end{table}

\begin{figure*}[htbp]
\centerline{\includegraphics[width=4in]{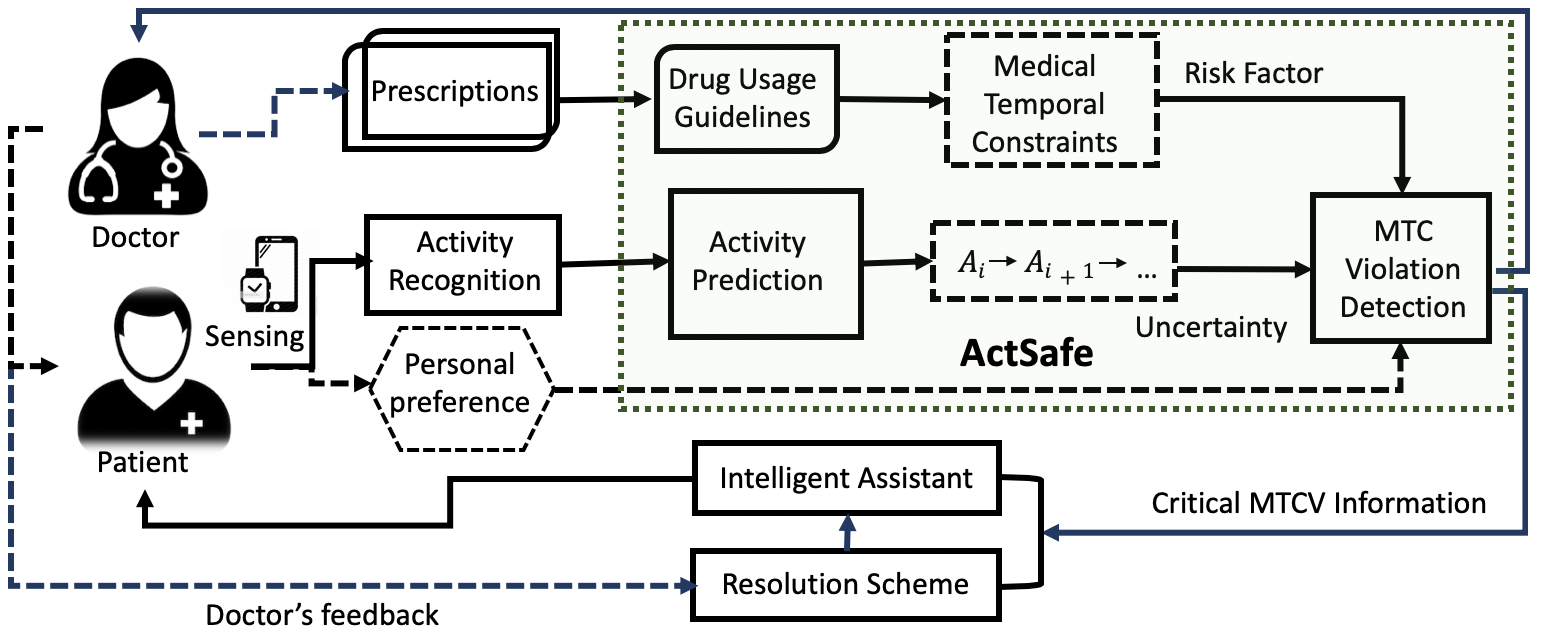}}
\caption{Overview of \paperNameActsafe{} and how it can be integrated into a decision support pipeline for patients and doctors. The rectangle with dotted-line border encloses the scope of \paperNameActsafe{}. \paperNameActsafe{} intercepts the regular health behaviors (RHBs) detected using an underlying activity recognition system. Next, it models the behaviors and predicts the next occurrence of a behavior. It also intercepts the drug usage guidelines for each prescription medication prescribed by the patient's doctors(s). Then it extracts medical temporal constraints (MTCs) from the set of advice that overlaps with RHBs. Finally, it identifies potential violations of MTCs between these two streams of data. This critical information can be forwarded to a patient through a personalized intelligent assistant.}
\label{system_diagram}
\end{figure*}
\section{Usage Scenario}
\label{sec_usage}
Before providing the solution outline we present an example scenario in which utilizing \paperNameActsafe{} would lead to better health outcomes in everyday living situations.


Kate is a 50-year-old female who was recently diagnosed with \textbf{arthritis}. She has also had an \textbf{atherosclerotic cardiovascular disease} since she was a teenager. To treat her atherosclerotic cardiovascular disease she has to take an ACE inhibitor at the same time each morning on an empty stomach. She has to be careful not to eat within 2 hours of taking her ACE inhibitor to ensure the medication is absorbed properly. To treat her arthritis, her physician has prescribed her the steroid prednisolone with her breakfast because if prednisolone is not taken with food or milk it can lead to stomach ulcers. Kate sometimes forgets to maintain the interval between these consecutive medications and ends up having breakfast shortly after taking her ACE inhibitor, mitigating its effects and putting herself at risk of having a severe build up of cholesterol in her arteries. Although she tried using a medication reminder app to manage her medications, the app is not aware of her daily behavior patterns, specifically what time she has eaten breakfast, and does not consider the dependencies between her medications and other daily behaviors. Thus, the reminders are hardly helpful. She often gets confused with all the temporal constraints her medications add to her daily basic behaviors, like sleeping or eating meals. Additionally, due to her confusion she is at risk of her medications not working properly leading to severe, negative health outcomes.

This scenario depicts several medical temporal constraints (MTCs), including (i) taking a medication (ACE inhibitor) on an empty stomach, (ii) taking a medication (ACE inhibitor) at the same time each day, (iii) taking a medication with breakfast (prednisolone). We present a set of example constraints from prescription medications for different chronic conditions in Table \ref{table_DUGAdvice}. Like Kate, many patients often suffer from confusion, frustration, and often ineffective treatment due to such complex temporal constraints their medications place on their everyday activities. While Kate takes two medications, patients with multiple chronic conditions may need to take many more medications, even 15-20 per day, each complicating their daily schedule \cite{ferguson2017barriers,dolce1991medication,donnan2002adherence,paes1997impact, ingersoll2008impact, mayer2001strategies, klakegg2018assisted}.

\paperNameActsafe{} can be used to assist patients such as Kate to safely adhere to their medication regimen. \paperNameActsafe{} extracts and models the medical temporal constraints (MTCs) belonging to each of her medications. It also considers how the MTCs interfere with regular health behaviors (RHBs). It uses a novel predictive modeling technique to predict her sequence of future behaviors based on her past behavior patterns. By modeling the MTCs from her prescription medications and predicting RHBs, \paperNameActsafe{} can predict potential violations of Kate's MTCs well ahead of time, giving her ample time to change course and follow her MTCs.

\section{Solution}

An overview of \paperNameActsafe{} and its usage flow are shown in \textbf{Figure \ref{system_diagram}}. \paperNameActsafe{} contains 4 components: (i) modeling MTCs in a generalizable way, (ii) extracting medical temporal constraints (MTCs) from free-format text and mapping them to the model, (iii) predicting RHBs related to MTCs, and (iv) predicting violations of MTCs. \paperNameActsafe{} extracts the medical temporal constraints (MTCs) from the prescribed drug usage guideline. \paperNameActsafe{} intercepts the behaviors recognized by an activity recognition engine. \paperNameActsafe{} can also be extended to intercept an individual's personal preferences (e.g., calendar) to sense intended future behaviors. By aligning the set of MTCs and potential future behavior sequences, \paperNameActsafe{}  predicts violations of MTCs. \paperNameActsafe{} can be used by physicians and patients together, and it can be integrated into an intelligent personal assistant (e.g., Alexa) to enhance the health safety and effectiveness of clinical treatment. 
The predicted violation of MTCs can be forwarded to these assistants in addition to the patient's care provider for potential resolution.

\subsection{Modeling Medical Temporal Constraints} \label{modeling_mtcs}
Medical temporal constraints (MTCs) can originate from doctor's suggestions, prescription medication labels, or personal preferences of the individual. Several examples of MTCs are shown in Table \ref{table_DUGAdvice}. 
In this paper, we mainly focus on MTCs originating from prescription medications related to chronic diseases. For any prescription medication, there exists a drug usage guideline (DUG) or patient handout \cite{Medscape} that contains critical information to ensure effective treatment, patient safety, and medication adherence. These guidelines or patient handouts are also available online. Often such information or advice contains MTCs, as shown in Table \ref{table_DUGAdvice}. Modeling medical temporal constraints (MTCs) from the unstructured, free-format text found in the drug usage guideline (DUG) poses the following challenges. (i) MTCs vary in terms of temporal precision: some MTCs are definitive, while some are imprecise. (ii) MTCs can arise either from a single behavior (e.g., taking medication \textit{n} times with \textit{m} hour intervals) or from dependencies between behavior pairs (e.g., taking medication \textit{m} hours before having a meal).

\subsubsection{Taxonomy of MTCs}
Based on our review, the MTCs originating from prescription medications can be either definitive or imprecise. Definitive MTCs can be further categorized into three classes: dependency, frequency, and interval. Imprecise MTCs can be further categorized into four classes: dependency, time dependency, consistency, and time-of-day. 

\begin{enumerate}

\item Definitive dependency constraints capture temporal dependencies between taking medication and other regular health behaviors (RHBs). For example, from the drug usage guideline (DUG) document of Protonix: "If you are taking the granules, take your dose \textbf{30 minutes before a meal}. " 


\item Frequency constraints capture the temporal constraints regarding the suggested frequency of a medication administration, i.e., how many times a medication should be taken in a specific interval (e.g., day, week). For example, from the DUG document of Wellbutrin: "Take this medication by mouth, with or without food, usually \textbf{three times daily}."


\item Interval constraints capture the temporal constraints regarding the suggested interval between consecutive medication administrations. For example, from the DUG document of Wellbutrin: "It is important to take your doses \textbf{at least 6 hours apart} or as directed by your doctor to decrease your risk of having a seizure."

\item Imprecise dependency constraints capture inexact temporal dependencies between taking medication and other regular health behaviors. For example, from the DUG document of Singulair: "Do not \textbf{take a dose before exercise} if you are already taking this medication daily for asthma or allergies. Doing so may increase the risk of side effects." Here the other related RHB is exercise.

\item Time dependency constraints capture inexact temporal dependencies between taking medication and a specific time of day. For example, from the DUG for the medication Prednisone: "If you are prescribed only one dose per day, take it in the morning \textbf{before 9 AM}." Here the dependency is imprecise ("before") and the time of day is 9 AM.

\item Consistency constraints capture the requirement to take a medication at the same time each day. For example, from the DUG document of Zocor: "Remember to take it at the \textbf{same time each day}."

\item Time-of-day constraints capture the requirement to take a medication at a certain time of a day. Take for example the DUG for the medication Prednisone: "If you are prescribed only one dose per day, take it \textbf{in the morning}."

\end{enumerate}

Multiple MTCs can appear in the same advice. For instance, consider the following statement from the DUG document of Starlix: "Take this medication by mouth \textbf{1-30 minutes before each main meal}, usually \textbf{3 times daily}, or as directed by your doctor." Here the statement has definitive dependency and frequency constraints.

\subsubsection{Modeling MTCs using Context-free-grammar}
A formal grammar is "context free" if its production rules can be applied regardless of the context of a nonterminal. The taxonomy of the MTCs mentioned above motivated us to develop a context-free-grammar based solution to model these definitive and imprecise MTCs. Here the non-terminals can be any medical temporal constraints (MTCs) and dependency among taking medication and other regular health behaviors. A context-free grammar is a suitable solution to model such dependencies and constraints as the production rule can be applied to any relevant datasets regardless of the context of the non-terminal, i.e., different types of MTCs. The novel grammar developed and integrated in \paperNameActsafe{} contains the following set of terminals.

\begin{itemize}
    
    \item natural number, $n$: 1 $\mid$ 2 $\mid$ 3...
    \item behavior, $act$: sleeping $\mid$ eating $\mid$ taking medication $\mid$ ...
    
    \item prepositions of temporal dependency, $dp$: before $\mid$ after 
    
    \item prepositions of interval dependency, $ip$: within $\mid$ for $\mid$ apart
    
    \item prepositions of occurrence, $p$: at $\mid$ in
    
    \item unit of time slots, $u$: hour $\mid$ minute $\mid$ day $\mid$ week
    
    \item time stamp, $t$: the same time $\mid$ 9 am $\mid$ 10.30 pm $\mid$ ...
    \item time of the day, $d$: morning $\mid$ evening $\mid$ noon
\end{itemize}

Using these terminals, the MTCs can be expressed using the following variables.

\begin{enumerate}

\item Definitive dependency constraint: $V_1$: $n$.$u$.$dp$.$act$ (e.g., 30 minutes before a meal)
\item Frequency constraint:  $V_2$: $n$ times in a $u$ (e.g., three times a day)
\item Interval constraint: $V_3$: $n$.$u$.$ip$ (e.g., 6 hours apart)
\item Imprecise dependency constraint: 
$V_4$: $dp$.$act$ (e.g., before meal)

\item Imprecise time dependency constraint: 
$V_5$: $dp$.$t$ (e.g., before 9 AM)

\item Consistency constraint: $V_6$: $p$.$t$ each $u$ (e.g., at the same time each day or at 9 am each day)

\item Time-of-day constraint: $V_7$: $p$.$d$ (e.g., in morning)

\end{enumerate}
The definitive dependency constraint and the imprecise dependency constraint can be merged into one dependency constraint. However, we distinguish between them to provide additional level of details while notifying patients about potential violations of MTCs. Also, the above-mentioned grammar can be used to model \textbf{compound} MTCs. For instance, taking a medication 2 hours before meal ($V_1$), 3 times a day ($V_2$), and 4 hours apart ($V_3$) can be expressed as: 
$V_i$: $V_1$.$V_2$.$V_3$. This grammar can also be extended to model \textbf{negated} MTCs. For instance, "\textit{do not take this medication before exercise}" can be modeled as, 
$\lnot$ $V_4$, where, $V_4$: $dp$.$act$ (e.g., before exercise).


\subsection{Extracting and Mapping Medical Temporal Constraints from Prescription Medications}
\label{MTC_extraction_solution}
To correctly predict MTC violations, MTCs must first be extracted from free-formatted patient education materials, such as drug usage guidelines. 
First, advice statements containing potential MTCs are extracted from the drug usage guideline (DUG) document of a prescription medication using a publicly available semi-automatic annotation tool \cite{preum2018corpus}. Alternatively, the patient can share the relevant text via their smartphone camera
by taking a picture of the drug usage label, or vocally via a virtual assistant, e.g. Amazon Alexa or Google
Assistant. The MTCs contained in the advice statements is then mapped to the context free grammar (CFG) based MTC model using the following strategy.

In the substring matching pipeline, we extract and map the relevant substring in a statement extracted from the drug usage guidelines to the list of MTCs. This utilizes the context free grammar (CFG) model proposed in Section \ref{modeling_mtcs}. First a list of all possible MTCs is enumerated for each MTC types. Each of the possible MTCs is tokenized. Since natural language frequently contains extraneous words that do not need to be extracted, each token is matched as a substring in the advice statement. If it is found, the next token is matched to the portion of the advice statement following the previously previously-matched token. This novel method is a simple yet effective solution for both matching advice statements with their containing MTC types and extracting those MTCs. As an example of this method, consider the real advice statement "If you also take certain other drugs to lower your cholesterol, ... take pravastatin at least 1 hour before or at least 4 hours after taking these medications." Both the "\textit{1 hour before taking medications}" and the "\textit{4 hours after taking medications}" CFG enumerations would be matched and extracted from this advice statement.



    

It should be noted that this step should be performed only once for a single patient unless they have a change in prescription or treatment. Also, we do not envision this step to be fully automated since the patient and caregiver should review the MTCs before they are used for predicting violations. Because MTCs are often subjective (i.e., different individuals might be affected differently by the same MTC) and their severity might vary based on physiological and environmental factors and patients' prognosis. 

\subsection{Predicting Regular Health Behaviors}
\label{ActPredSolution}
Existing works propose alternative activity prediction models, where each model is suitable for different datasets and applications \cite{minor2015data}. Some works report traditional ML models perform the best \cite{minor2015data}, while some works report deep learning models to outperform traditional statistical ML models \cite{kim2017activity}. However, the following factors limit the applicability of existing activity prediction solutions for \paperNameActsafe{}. 

\begin{itemize}
    
    
    \item The RHBs related to MTCs vary drastically in terms of average duration of behaviors: average duration of \textit{sleeping} can be a few hours where the average duration of \textit{taking medication} is often less than a minute as observed in public datasets \cite{cook2009assessing}. Also, some relevant behaviors have dynamic duration, e.g., \textit{eating}. This issue is even more challenging for an unbalanced dataset, where the related health behaviors have much fewer samples than other behaviors. Most of the existing activity prediction solutions do not consider such short-span, less-frequent, yet critical behaviors or behaviors with dynamic span.
    
    \item Most existing activity prediction solutions accurately predict the start of an activity with a short prediction horizon, e.g., predicting a few minutes ahead, or they can predict the next activity \cite{minor2015data, preum2015maper, minor2017forecasting, krishna2018lstm, yang2020multi}. They do not report the effectiveness of the model over a long prediction horizon (e.g., 12 hours, or 8 hours). However, since the end goal of \paperNameActsafe{} is to predict MTCs well ahead of time to provide patients and healthcare providers with enough time to correct the violation, the behavior prediction model should be robust enough to predict with reasonable accuracy over a long time span.

\end{itemize}

\begin{figure*}[htbp]
\centerline{\includegraphics[width=3.5in]{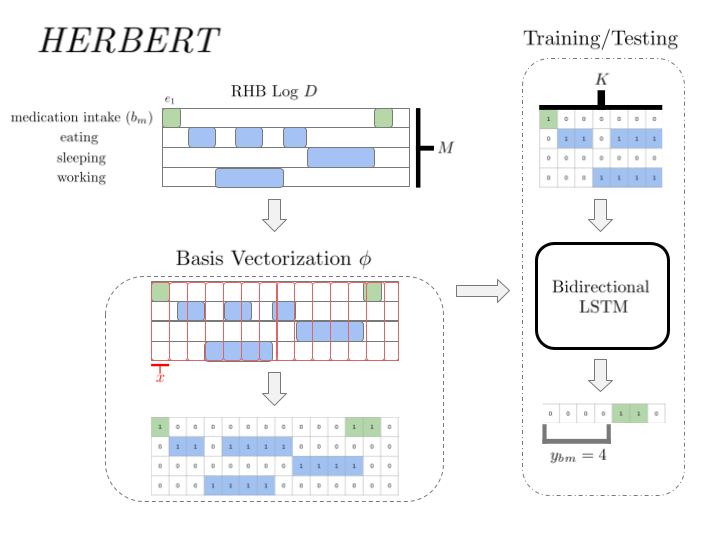}}
\caption{Overview of the \fancyModelName{} model architecture. The \fancyModelName{} utilizes the basis vectorization function described in Algorithm \ref{basis_vectorization_algorithm} to discretize entries of an RHB log into $x$-minute windows. It then trains a bidirectional LSTM network to predict the number of windows, $y_{b_i}$, until the next occurrence of a target activity $b_i$.}
\label{herbert_diagram}
\end{figure*}

We address these challenges by proposing the HEalth Related BEhavior pRedicTion (\fancyModelName{}) model. \fancyModelName{} formulates the RHB prediction problem as a regression timestamp prediction problem utilizing basis vectorized prior behavior sequences, where occurrence times of each RHB are predicted independently over a robust prediction horizon \cite{preum2015maper}. The RHBs predicted by the \fancyModelName{} model are a small set of RHBs performed by a patient such as \textit{medication intake}, \textit{eating}, and \textit{sleep}. Critically, the \fancyModelName{} model utilizes high-level behavior/activity labels generated from the sensor readings of a human activity recognition system. Thus it relies only on previously recognized RHB timestamps and their durations to predict future timestamps of a behavior/RHB. Specifically, consider a set of RHBs $B = \{b_1, \ldots, b_M\}$. To construct an \textit{RHB entry} $e$ for a specific RHB $b_i \in B$, we use a tuple $e = (b_i, t_{start}, t_{stop})$, where $t_{start}$ is the timestamp at which the patient started RHB $b_i$ and $t_{stop}$ is the timestamp at which the patient stopped that RHB. An \textit{RHB Log} $D$ of size  $|D| = Q$ is a set of RHB entries ordered by start time, with $e_1 = (b_{i_1}, t_{start_1}, t_{stop_1})$ being the first RHB recorded by the patient during the study and $e_{Q} = (b_{j_{Q}}, t_{start_{Q}}, t_{stop_{Q}})$ being the final RHB recorded by the patient during the study. We define \textit{Basis vectorization} to be a function $\phi: (D,B,x) \xrightarrow{} \mathbb{R}^{M \times K}$ where $M = |B|$ and $K$ is the number of $x$-minute windows between $t_{start_1}$ and $t_{stop_{Q}}$, which defines a matrix $A \in \mathbb{R}^{M \times K}$. Each element $a_{i,j} \in A$ is either a $1$ if behavior $i$ was recorded in the $j$th window, or $0$ otherwise. A detailed basis vectorization algorithm is given in Section \ref{appendix_basis_vec}. Then for a target RHB $b_i \in B$, \fancyModelName{} predicts the number of $x$-minute windows until the next occurrences of $b_i$, $y_{b_i}$, by utilizing the prior $T$ weeks of RHBs formatted as basis vectors. For example, setting $T=3$ for 3 weeks' worth of RHBs and $x=30$ for 30-minute time intervals, we have $K = \frac{3 \cdot 7 \cdot 24 \cdot 60}{30}$ (in minutes) $= 1008$ windows, so $K=1008$. The \fancyModelName{} model for $b_i$ then predicts the number of $x$-minute windows $y_{b_i}$ until the next occurrence of the target RHB $b_i$. We include the basis vectorization algorithm in Algorithm \ref{basis_vectorization_algorithm} in Section \ref{appendix_basis_vec}, and a diagram of \fancyModelName{} in Figure \ref{herbert_diagram}.

We model \fancyModelName{} to utilize the basis vectorization process for two reasons. First, basis vectorization smooths sensor labels which can often be erratic in nature.
By considering only whether or not that labeled RHB occurred during a given window, basis vectorization smooths the noisy data. Second, basis vectorization mitigates the differences between short-span RHBs such as \textit{medication intake} and long-span RHBs such as \textit{sleeping} by giving each a larger unit of time (e.g. a $30$-minute window). We also utilize \fancyModelName{} to predict the timestamp of the next RHB occurrence. We make this design choice so \fancyModelName{} can account for longer prediction horizons. We choose to utilize a non-parametric deep learning-based model in \fancyModelName{} for the prediction task to eliminate feature engineering. \fancyModelName{} utilizes a Long-Short-Term Memory (LSTM) deep learning architecture over with up to 3 weeks' worth of basis vectors (i.e. $K=2016$ windows when $x=15$) to predict RHB occurrences. Casagrande et al. \cite{casagrande2019sensor} showed that it is possible to achieve acceptable prediction accuracy with up to three weeks of collected data, motivating the use of the 3-week activity basis vectorization window utilized by the \fancyModelName{} model \cite{casagrande2019sensor}. Hence the \fancyModelName{} model implicitly utilizes the following two features for prediction.

\begin{enumerate}
    \item Behavior lag: The lag of the target behavior.
    \item Interaction among multiple behaviors: Often, one behavior impacts the occurrence and duration of another behavior. For example, if someone wakes up later, the morning medication time is also likely to be delayed.
\end{enumerate}

\subsection{Predicting Violations of MTCs}
We predict violations of MTCs by developing a new rule base that aligns the semantics of behavior sequence and temporal constraints of medical advice to predict potential violations well ahead of the possible time of violations. This rule base consists of around 30 rules to detect violations of each type of MTC from a predicted sequences of behaviors. This rule base is extensible to other datasets, other regular health related behaviors, and medical temporal constraints. 
For instance, to predict consistency violation, one rule compares the timestamp of a predicted medication intake RHB with a predefined timestamp indicating the suggested time to take the medication, allowing a flexible window of 15 minutes before and after the timestamp. If the predicted timestamp does not fall within the 15-minute window, then the rule flags the predicted task of taking medication as a violation of the consistency constraint. When predicting precise dependency violations, e.g. "2 hours before breakfast", our rule base utilizes the predicted timestamp of the \textit{breakfast} RHB and the predicted timestamp of the \textit{medication intake} RHB. If the \textit{medication intake} predicted timestamp falls within the precise dependency, in this case within the 2 hours before the predicted \textit{breakfast} timestamp, a violation is detected.

\begin{table}
\caption{\textbf{MTC Distribution}: The distribution of different types MTCs observed in the prescription dataset. The last column shows the count of different MTCs found in the chronic disease prescription dataset.}
\label{table_CountTCVDataset}
\begin{tabular}{ccccc}
\toprule
Precision                   & Type  & Name of MTC     & Example of MTC           & MTC Count \\ \midrule
\multirow{3}{*}{Definitive} & MTC 1 & Dependency      & 2 hours before breakfast & 52        \\
                            & MTC 2 & Frequency       & 2 times a day            & 32        \\
                            & MTC 3 & Interval        & 4 hours apart            & 6         \\ \midrule
\multirow{4}{*}{Imprecise}  & MTC 4 & Dependency      & after eating             & 53        \\
                            & MTC 5 & Time Dependency & before 9 a.m.            & 1         \\
                            & MTC 6 & Consistency     & same time each day       & 95        \\
                            & MTC 7 & Time-of-day     & evening                  & 16        \\ \bottomrule
\end{tabular}
\end{table}

\section{Experimental Setting}
\paperNameActsafe{} can be applied to improve adherence to any medications or treatment option with temporal constraints. Because managing chronic diseases often impose MTCs on patients, in this paper we evaluate \paperNameActsafe{} in the context of chronic diseases. We evaluate the performance of the \textit{MTC extraction solution} presented Section \ref{MTC_extraction_solution} using a real prescription dataset for chronic diseases as described below. We evaluate the performance of predicting regular health behaviors (RHB) and violations of MTCs using a real RHB dataset collected from $N=28$ chronic disease patients in a real-world uncontrolled setting.


\subsection{Prescription Dataset}
The prescription dataset contains 35 real prescriptions from patients with multiple chronic diseases \cite{mtsamples}. This dataset is annotated with MTCs assigned to each of the prescription medications. This dataset is utilized to demonstrate the coverage of the MTC taxonomy introduced in Section \ref{modeling_mtcs}. This dataset contains 83 prescription medications from several chronic diseases, including but not limited to diabetes mellitus (type I and type II), bipolar affective disorder, depression, hypertension, hypotension, hyperlipidemia, hypothyroidism, bradycardia, chronic pain, morbid obesity, osteoarthritis, and obstructive sleep apnea. For each of these medications, the corresponding drug user guideline (DUG) is extracted from a DUG corpus \cite{preum2018corpus}. Each DUG contains advice to guide patients to take the medication properly and safely which are manually annotated by two annotators to identify and categorize the medical temporal constraints. Sample advice statements are presented in Table \ref{table_DUGAdvice}. 

The distribution of different MTCs observed in the prescription dataset used for evaluating \paperNameActsafe{} are presented in Table \ref{table_CountTCVDataset}. This is calculated as follows. For each medication in a prescription, relevant advice statements from the DUG document of the medication are extracted. Those statements are manually annotated to indicate which MTC types they contain. Table \ref{table_CountTCVDataset} contains the aggregated count of MTCs across the dataset of 35 prescriptions in the fourth column. Here, the most common type of MTCs are \textit{consistency} with a frequency of 95. 

\subsection{Regular Health Behavior (RHB) Dataset}
\label{RHB_dataset}
To demonstrate the effectiveness of \fancyModelName{} in a real-world scenario, we utilize a real RHB dataset. This dataset contains data from $n=40$ patients diagnosed with atherosclerotic cardiovascular disease (ACD) and prescribed to take a daily medication namely ACE inhibitors to manage their condition. In a study between October 2018 and April 2019, daily smartphone sensor data was recorded from each of 40 participants' Android smartphones. These 40 participants were recruited for an eight-week study. Of the 40 participants in the study, 7 had less than one week of medication intake data (not enough data to train \fancyModelName{}), 4 had unusable time formatting in the logged data, and 1 had no RHB labels outside of medication intake. Data from these 12 participants were deemed unsuitable for evaluating \fancyModelName{}, hence we use data from the remaining $N=28$ participants.

The study collected medication intake and other RHB data for 8 weeks in an uncontrolled environment, i.e. throughout the participants' regular daily life. Sensor readings were first labeled by the patient declaring which behavior they were participating in when prompted, then an active model framework was used to learn and record behavior labels for sensor readings throughout the rest of the study period. Both the model and participant behavior labels were considered as ground truth RHB logs in our evaluation. The \fancyModelName{} model utilizes only the participants' RHB logs, not individual sensor readings. In addition to the Android smartphone which tracked the participants' activities and created the RHB dataset, the participants were given Bluetooth-enabled pill bottles which recorded time(s) at which they took medication each day. Each patient had a prescription for an \textbf{ACE inhibitor} which was to be \textbf{taken on an empty stomach }at the \textbf{same time each day}.

Each patient RHB log differs slightly in length. On average each patient has 52.65 days worth of RHB log data. The average \textbf{training set size} for each patient was 37.79 days and the average \textbf{test set size} was 14.86 days. Details about patient demographics in the RHB dataset and a sample of one of these logs are provided in Section \ref{appendix_rhb_log}, and a sample of patient-level data size statistics is given in Section \ref{appendix_tables}. We also take in to account the regularity of behavior and sparsity / missingness of data to capture real-world challenges of behavior prediction in Section \ref{regularity_sparsity_results}. 

This dataset is suitable for our evaluation of the \fancyModelName{} model and the  \paperNameActsafe{} framework. The RHB logs are passively collected long-term, in real-world settings for chronic disease patients using Android and Bluetooth devices. The eight-week monitoring period is sufficient to capture behavior variation across multiple days, weeks, and months. The data represent real RHB patterns of real chronic disease patients, along with noise and variability that occur naturally in a realistic uncontrolled environment.

\section{Results}
Our evaluation of the \paperNameActsafe{} framework consists of \textbf{three} parts. \textbf{First}, in Section \ref{extracting_medical_temporal_constraints_results}, we test the capacity of the substring matching pipeline for MTC extraction from free-form drug usage guidelines in the prescription dataset by framing the problem as a multi-class classification problem. \textbf{Second}, we investigate the behavior prediction model \fancyModelName{} in Section \ref{behavior_prediction_results}. We compare the \fancyModelName{} model with several baseline models when predicting the medication intake RHBs for each of the patients in the RHB dataset. We investigate the effects of $x$-minute window size, schedule regularity, and sparsity on model performance.  \textbf{Finally}, in Section \ref{predicting_violation_results_section} we evaluate the performance of \paperNameActsafe{} when predicting MTC violations for real chronic diseases in the RHB dataset.

\subsection{Extracting Medical Temporal Constraints}
\label{extracting_medical_temporal_constraints_results}

To test the effectiveness of our substring matching algorithm, we attempt to extract each type of MTC from the samples in the prescription dataset and compare our results with ground truth. We treat each MTC match as a correct classification if the MTC types are correctly predicted from the free-form drug usage guidelines for a given medication. The substring matching classification results are displayed in Table \ref{mtc_extraction_table_substring}. We report standard precision, recall and F1-score. We also list the count of occurrences of each type of MTCs.

We could not evaluate the performance of our pipeline for MTC 3 and MTC 5 due to very limited number of samples in the prescription dataset used for evaluation. The substring matching results are promising for our dataset, scoring a 0.96 F1 score or higher across the 5 MTC types evaluated in this experiment. However, there are a few instances in our drug usage guideline dataset that mention MTCs implicitly and thus the corresponding MTCs cannot be extracted. As an example, the drug usage guideline for the drug Ismo states: \textit{take this medication by mouth a directed by your doctor, usually once daily when you wake up}. This statement contains a type 4 MTC, i.e., "after waking up" as implied by "daily when you wake up". Although this MTC is obvious to a human reader, it does not match the \paperNameActsafe{} CFG and is overlooked by the substring matching algorithm.

\begin{table}[h]
\caption{\textbf{MTC Extraction Results}: Performance of the substring matching algorithm within the MTC extraction pipeline. In this classification task, a correctly-extracted MTC is considered a correct classification. The substring matching algorithm  successfully extracts each of the 5 types of MTCs from the drug usage guideline dataset with an average F1 score of 0.98. MTC types 3 and 5 could not be evaluated in this experiment due to lack of instances in the prescription dataset.}
\label{mtc_extraction_table_substring}
\begin{tabular}{ccccccc}
\toprule
MTC Type & Precision & Recall & F1   & Count \\ \midrule
MTC 1   & 0.98      & 0.93   & 0.96 & 22    \\
MTC 2   & 0.97      & 1.00   & 0.98 & 17    \\
MTC 4   & 0.98      & 0.97   & 0.98 & 29    \\
MTC 6   & 0.98      & 1.00   & 0.99 & 45    \\
MTC 7   & 0.97      & 1.00   & 0.98 & 22    \\ \bottomrule
\end{tabular}
\end{table}

\subsection{Behavior Prediction}
\label{behavior_prediction_results}

We use root-mean-square error (RMSE) as the performance metric. The output of \fancyModelName{} is $\hat{y}_{b_i} \in \mathbb{R}$, the predicted number of windows until the next occurrence of RHB $b_i$. The true number of windows until the next occurrence of RHB $b_i$ is $y_{b_i}$. When predictions over test frames are compared against ground truth, we use RMSE to measure the total distance from the predictions to the ground truth.


\subsubsection{Effect of Varying Window Sizes on \fancyModelName{} and Baselines}
\label{medintake_baselines}
We first compare the performance of \fancyModelName{} with baseline models across $3$ different $x$-minute window sizes: $x \in \{15, 30, 60\}$. We choose these three window sizes because RHBs are often scheduled as 15-minute, 30-minute, or 1-hour activities, or some combination of these time segments. Since previous approaches to activity prediction are all dependent on prior sensor readings to predict the timestamp of a given activity \cite{hao2018activity, ghods2019activity2vec, casagrande2019sensor} and thus rely on differing inputs, comparing the performance of \fancyModelName{} with these models is not fair. Hence, we consider only statistical models as our baselines for time series prediction. We describe and motivate two baseline models below. 

\begin{enumerate}
    \item \textit{Prior-day baseline}: Since RHBs are often repeated daily at scheduled times, a naive approach to human behavior prediction would be to simply assume that yesterday's behaviors are a perfect predictor of today's behaviors. Hence for any given RHB $a$, our prior-day baseline simply takes the previous time of day at which $a$ was performed and predicts that time of day for the coming day. This approach is agnostic to any temporal interaction between RHBs and can be used to compare the effect \fancyModelName{}'s use of RHB basis vector embedding for improving prediction performance. 
    
    \item \textit{ARIMA}: The ARIMA time series model is a context-free statistical model that considers only the lag of the predicted value \cite{arima}. As ARIMA considers only the basis vectors of the prior occurrences of the predicted RHB $b_i$, using this model as a baseline to predict the \textit{medication intake} RHB and other RHBs will allow us to determine if related RHB context is useful in predicting future health behaviors.
    
    \item  \textit{LSTM}: A Long-Short-Term Memory (LSTM) deep learning architecture without the basis vectorization process. In this model we train an LSTM to predict the timestamp (in seconds) of the next target RHB $b_i$ by utilizing the prior $25$ RHB entries in a patient behavior log $D$. We include this baseline to measure the effectiveness of basis vectorization within the \fancyModelName{} architecture. Since \fancyModelName{} predicts $y_{b_i}$, the number of $x$-minute windows until the next occurrence of $b_i$, RMSE for the LSTM baseline model is normalized to number of $x$-minute windows for comparison.
\end{enumerate}

In Table \ref{baseline_results_table} we record model RMSE results for predicting the \textit{medication intake} ($m$) RHB, hereafter referred to as $b_{m}$. The RMSE results in this table are averaged across each sample in the test set of for each of the patients in the RHB dataset. We note than since each model except the LSTM predicts $y_{b_{m}}$, the number of windows until the next occurrence of the \textit{take medicine} RHB, the predicted value is larger for smaller window sizes (i.e., one $60$-minute window is equal to two $30$-minute windows and four $15$-minute windows). To obtain a single performance measurement for \fancyModelName{} we take average RMSE across all RHB patients for each model, then take the mean percentage decrease in \fancyModelName{} RMSE compared with the three baseline models RMSE. In the $30$-minute setting \fancyModelName{} outperforms baselines by an average of 57\%, with average percentage RMSE decreases of 42\% and 54\% in the $15$ and $30$ minute window size settings, respectively. Averaging these percentages across the three window size settings we see that \fancyModelName{} outperforms baselines by 51\%. As schedules, sensor labeling, and patient preferences introduce varying degrees of noise between patients, the best-performing model may vary between patients\footnote{See Section \ref{appendix_tables} for a sample table of patient-level results.} in the RHB dataset. Averaging across patients, however, the best model in terms of RMSE (normalizing for window size) is the $30$-minute \fancyModelName{} model with an RMSE of $38.4015$. Hence we set the basis vectorization window size parameter $x$ to be $30$ in the rest of the evaluation. We also note that while \fancyModelName{} outperforms the three baselines in the $x=30$ and $x=60$ window size settings, it fails to beat either statistical baseline in the $x=15$-minute setting. We discuss a possible reason for this anomaly in the next section.

\begin{table}
\caption{\textbf{Baseline and $x$-minute window size comparison}: This table displays the average root mean squared error (RMSE) results for each of the baselines compared with the \fancyModelName{} model when predicting the \textit{medicine intake} RHB in the test sets of each of the patients in the RHB dataset. Since each model predicts $y_{b_{m}}$, the number of windows until the next occurrence of the \textit{take medicine} RHB, the predicted value is larger for smaller window sizes (i.e. one $60$-minute window is equal to two $30$-minute windows and four $15$-minute windows). When averaging across patients in the RHB dataset and window sizes, the \fancyModelName{} model outperforms the baselines by 51\%. The best model in terms of RMSE (normalizing for window size) is the $30$-minute \fancyModelName{} model.}
\label{baseline_results_table}
\begin{tabular}{lrrrr}\toprule
$x$-minute &Prior-Day &ARIMA &LSTM &HERBERT \\\midrule
15 &95.6782 &\textbf{90.4982} &314.0572 &96.0786 \\
30 &58.5409 &53.9906 &157.0286 &\textbf{38.4015} \\
60 &23.8988 &23.2102 &78.5143 &\textbf{19.4733} \\
\bottomrule
\end{tabular}
\end{table}

\subsubsection{Effect of Regularity of Schedule and Sparsity of Labelled Data}
\label{regularity_sparsity_results}
In this section we investigate two factors which possibly contribute to variation in \fancyModelName{} model performance across different RHB patients. Participants in the RHB dataset each took the same medication, an ACE inhibitor. However since schedules were recorded for each participant in an uncontrolled setting, each schedule was subject to some amount of noise and variability. Possible sources of noise and variation include but is not limited to (i) the time of day at which a participant/patient chose to take their medication, (ii) the degree to which they missed their medication, (iii) sensor-level noise , and (iv) normal day-to-day erratic scheduling throughout the study. These irregularity and sparse / missing labels make the prediction of medication intake and other RHBs more challenging. So we explicitly investigate the performance of \fancyModelName{} across different degrees of regularity of schedule and spareness/missingness of data for each participant in our RHB dataset. 

First, we examine \textbf{schedule regularity}. Since \textit{medication intake} is the most critical RHB in the \paperNameActsafe{} framework, we are particularly interested in how regularly a patient takes their medication. We specifically investigate how patient medication schedules differ compared to other patients in the RHB dataset, hence we define inter-patient regularity. To measure patient-level differences in \textit{medication intake} ($m$) schedule regularity, for a given patient $P$ we define a medication schedule $S_P \in \mathbb{R}^l$ to be the set of differences between the ordered occurrences of the RHB $b_{m}$. In other words, each element $s_i \in S_P$ is the time (in number of $30$-mnute windows) between $t_{stop_i}$ and $t_{start_{i+1}}$ for each $e_i = (b_{{m}_i}, t_{start_i}, t_{stop_i})$, $i \in \{0, \ldots, l-1\}$ with $l$ equal to the number of occurrences of the RHB $b_{m}$. We define $S_P$ in this way because it allows us to measure both how consistently a given patient timed their medication intake each day, as well as how frequently they missed or skipped a day of their medication intake throughout the study. To measure how similarly two patients $P_1$ and $P_2$ took their medication, we can use the cosine similarity function between their medication schedules, $sim(S_{P_1}, S_{P_2})$. Examination of pairwise medication schedule similarity is included in Section \ref{appendix_schedule_regularity}.

\begin{figure*}[htbp]
\caption{\fancyModelName{} model performance by medication intake schedule regularity across each RHB dataset in the $30$-minute basis vectorization setting. Regularity here refers, for patient $i$, to the sum of similarity between the patient's medication intake schedule $S_{P_i}$ and each other RHB patient medication intake schedule in the RHB dataset. We notice that, on average, less regular medication schedules led to worse model performance in terms of \fancyModelName{} model performance.}
\label{regularity_figure}
\centerline{\includegraphics[width=4in]{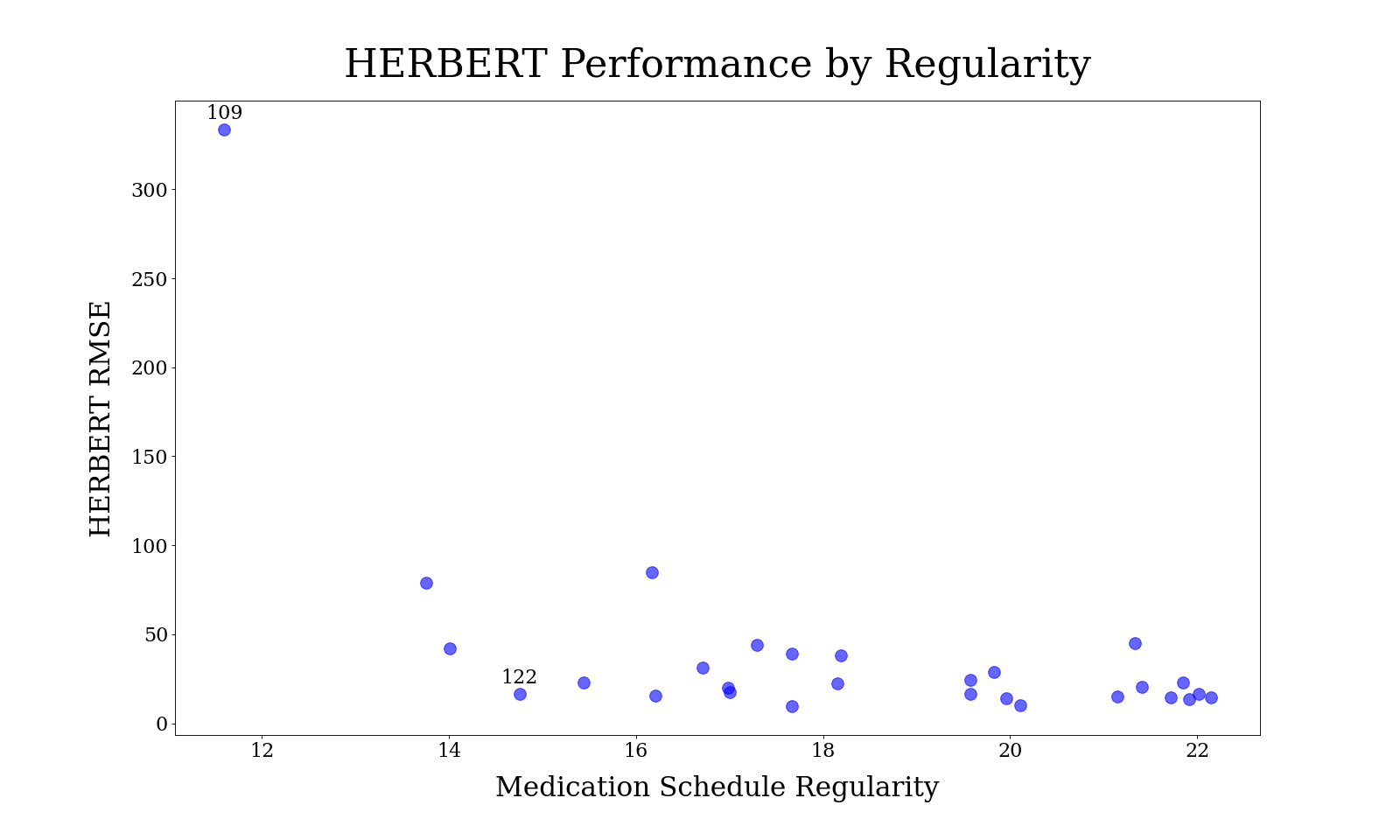}}
\end{figure*}

To measure how much patient $P_i$'s medication schedule differs from the whole distribution of patients in the RHB dataset, we simply take $reg(P_i)$ as the sum of cosine similarities between $S_{P_i}$ and $S_{P_j}$, i.e. $reg(P_i) = \sum\limits_{j \neq i} 1 - sim(S_{P_i}, S_{P_j})$. We call this value the \textit{regularity} of a schedule $S_{P_i}$, and it follows intuitively that a lower regularity for a patient $P$ indicates that their medication intake schedule was more erratic, i.e. if a patient has inconsistent gaps between medication intake entries their medication intake is more erratic. We see in Figure \ref{regularity_figure} the general trend that more irregular schedules result in worse \fancyModelName{} model performance. In particular, patient 109 had by far the most irregular medication intake schedule of any of the patients in the RHB dataset, and the \fancyModelName{} model performed by far the most poorly when predicting their medication intake RHB. Interestingly, while regularity seems to generally be correlated with model performance, \fancyModelName{} is robust to some amount of irregularity. User 122 had a low regularity score of just 14.75, but \fancyModelName{} performed well when predicting patient 122's medication intake RHB, with a RMSE of 16.67. There are many reasons a patient might have had a unique schedule over the course of the study. While all ACE inhibitors are subject to the same consistency constraint MTC, "same time(s) each day," this MTC may have been more challenging for certain patients leading to less regular medication schedules. Users could forget, have other non-RHB  activities get in the way, or decide to change schedules. In the context of \paperNameActsafe{}, a patient changing schedules would be recognizable. For example consecutive violations of the "same time(s) each day" consistency constrain MTC would be logged, and the \paperNameActsafe{} framework would be able to prompt the patient to ask if there has been a schedule change and ask them to update their framework parameters accordingly. Unfortunately, however, the reasons for medication schedule irregularity in the RHB dataset are unknown and thus not studied explicitly in this evaluation. Instead we focus on general medication schedule regularity.

\begin{figure*}[htbp]
\caption{Dataset sparsity in each of the $\{15, 30, 60\}$ minute basis vectorization settings. Sparsity here is measured as the percentage of cells in $BV_P$ which are $0$. Average sparsity and performance for each $x$-minute setting are also included. We notice that for smaller window sizes, sparsity increases. In the $15$-minute basis vector setting specifically, average sparsity is above 90\%. We hypothesize that to some degree, sparsity is influential in worse \fancyModelName{} test results in the $15$-minute basis vector setting.}
\label{sparsity_figure}
\centerline{\includegraphics[width=3.5in]{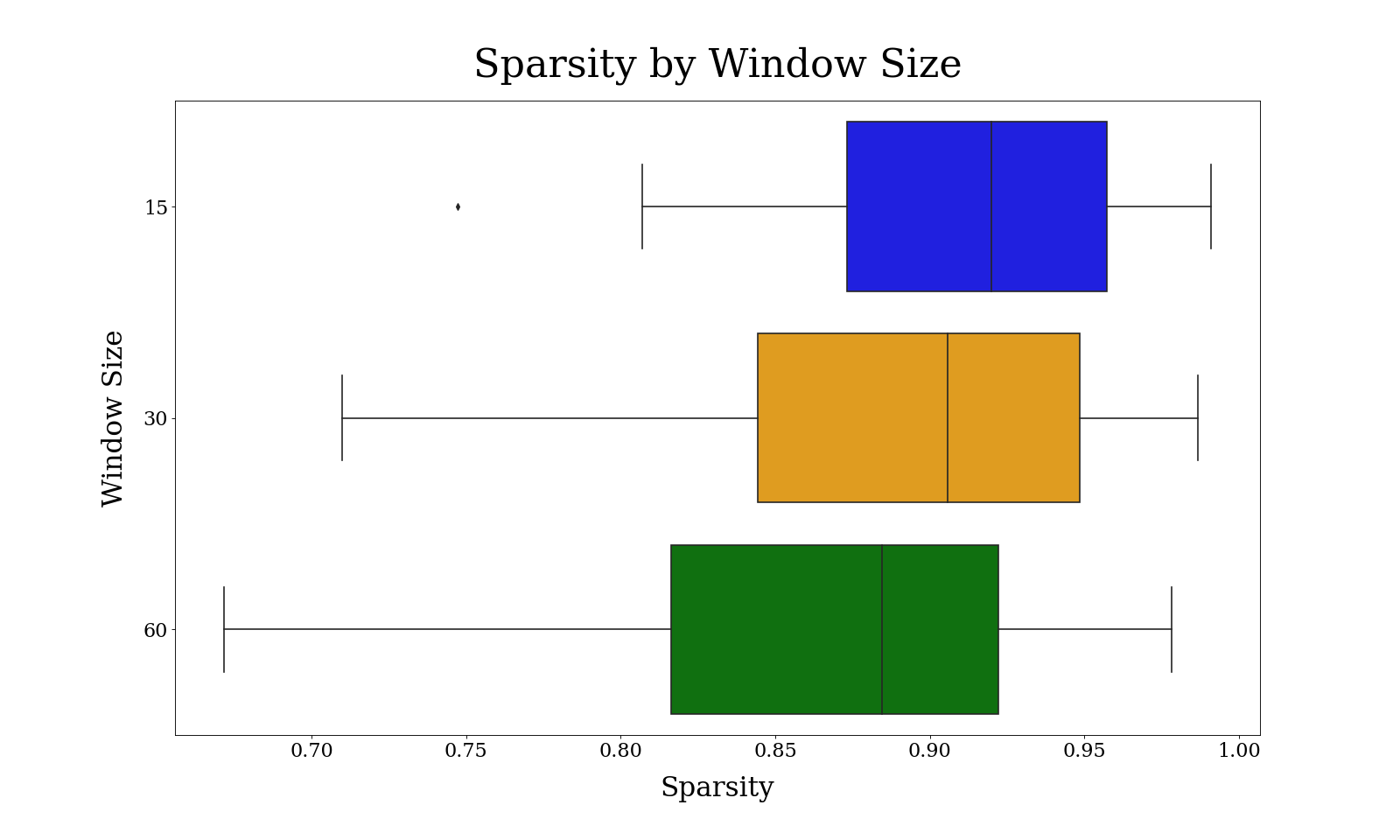}}
\end{figure*}

Next we examine \textbf{sparsity}. Deep learning models such as \fancyModelName{} are known to struggle in sparse data settings, where historical data contains large gaps \cite{eldesokey2018propagating}. These gaps challenge deep learning methods to generalize over large input spaces. Sparsity in the context of \paperNameActsafe{} could be due to many different reasons. For example, a patient not actually performing a regularly-scheduled RHB such as taking their medication would lead to sparsity. Perhaps an underlying human activity recognition system fails to record important RHBs, or a patient leaves their sensor-enabled smart device while on vacation. While we don't investigate ground truth as to why data sparsity occurs, we investigate how the \fancyModelName{} model performs under different degrees of missingness.

As defined by the basis vectorization function $\phi(D, x)$, a patient $P$'s schedule is a matrix $BV_P \in \mathbb{R}^{MxK}$. We define the sparsity of the matrix, $sparse(BV_P)$ to be the percentage of cells in $BV_P$ which are zeros. By definition of basis vectorization and sparsity, this will lead to an inverse relationship between window size $x$ and $sparse(BV_P)$, e.g., a $15$-minute basis vectorization will yield a larger, sparser matrix than a $60$-minute basis vectorization. In Figure \ref{sparsity_figure} we examine the relationship between sparsity and $x$-minute window sizes. We see that the smaller the basis vectorization window size parameter $x$, the higher sparsity becomes. In Section \ref{medintake_baselines} we've seen that the $15$-minute vectorizations have worse performance on average in terms of \fancyModelName{} RMSE. We hypothesize this decreased \fancyModelName{} model performance in $15$-minute vectors could be due to a combination of higher data sparsity and increased model size and complexity.

\subsection{Predicting Violations of Medical Temporal Constraints (MTCs)}
\label{predicting_violation_results_section}
We now evaluate the performance of \paperNameActsafe{} for the prediction of real chronic disease MTC violations across patients in the RHB dataset. As described in Section \ref{RHB_dataset}, our RHB dataset consisting of $N=28$ patients is adapted from a study which collected data from patients at risk for atherosclerotic cardiovascular disease. Each patient had a prescription for an ACE inhibitor, a class of drug which is typically associated with three major MTCs:

\begin{enumerate}
    \item "same time each day", a consistency constraint MTC (type 6)
    \item "2 hours before eating", a definitive dependency constraint MTC (type 1)
    \item "2 hours after eating", a definitive dependency constraint MTC (type 1)
\end{enumerate}

The second and third MTCs here are adapted from the DUG instruction "take [the ACE inhibitor] on an empty stomach." The National Institute of Aging defines taking medication on an empty stomach to mean "at least two hours before you eat or two hours after you eat." \cite{nioa} While in our analysis we assumed that a 2-hour gap between the \textit{eating} RHB and the \textit{medication intake} RHB represented this instruction, the time gap is a parameter that could be discussed between patient and physician and specially designated in the \paperNameActsafe{} framework.

\begin{table}
\caption{MTC violation prediction results comparing \fancyModelName{} and the best baseline model, ARIMA. We report weighted average precision, recall, and F1 for each of the two types of MTCs associated with the ACE inhibitors taken by patients in the RHB dataset. For each of the time frames in the test sets of each of the patients in the RHB dataset, a violation either occurs (1) if the MTC is violated during that time frame or it does not (0). Since there are two related definitive dependency constraints associated with ACE inhibitors, "2 hours before eating" and "2 hours after eating," we determine that a violation occurs if either constraint is violated. Each model's predicted timestamps are utilized by our novel MTC violation prediction rule base to predict whether a violation of that type will occur. \fancyModelName{} significantly outperforms ARIMA when predicting dependency constraints. \fancyModelName{} also outperforms ARIMA when predicting violations of the "same time each day" consistency constraint, though both models perform relatively well.}
\label{violation_results_table}
\begin{tabular}{lrrrrr}\toprule
MTC Type &Model &Precision &Recall &F1 \\\midrule
Definitive Dependency Constraint &\fancyModelName{} &\textbf{0.7201} &\textbf{0.7902} &\textbf{0.7442} \\
Definitive Dependency Constraint &ARIMA &0.0592 &0.1729 &0.0815 \\
Consistency Constraint &\fancyModelName{} &\textbf{0.9932} &\textbf{0.9845} &\textbf{0.9840} \\
Consistency Constraint &ARIMA &0.9911 &0.9792 &0.9817 \\
\bottomrule
\end{tabular}
\end{table}

Each of the patients in the RHB dataset was subject to these MTCs when taking their ACE inhibitor. Each RHB dataset consists of multiple test windows, ranging from 48 hours to slightly over 49 days, which were held out from the training data when training the \fancyModelName{} model. To evaluate the performance of \paperNameActsafe{} for the prediction of these $3$ ACE inhibitor MTCs, in each of the test frames for each of the patients in the RHB dataset we consider the \fancyModelName{} model's predicted timestamp for both the \textit{medication intake} RHB and the \textit{eating} RHB\footnote{While each of the $N=28$ RHB patients tracked their \textit{medication intake} RHB, $2$ patients did not label the \textit{eating} RHB, and are excluded from the analysis of the two definitive dependency constraint MTCs.}. Using our novel rule base capable of predicting five types of MTCs, \paperNameActsafe{} predicts whether an MTC violation will occur during that test frame. Using the actual timestamps of the next \textit{medication intake} RHB and the next \textit{eating} RHB, we determine if an MTC violation occurred during the test frame. Hence, we evaluate the task of predicting MTC violations as a binary classification task in which, for each MTC associated with ACE inhibitor intake, a given test frame either violates that MTC ($1$) or it does not ($0$). Treating each test frame of each RHB dataset separately, we have $10,912$ possible consistency constraint MTC violations and $9,344$ possible definitive dependency constraint MTC violations. Since the goal of \paperNameActsafe{} is twofold, to correctly predict MTC violations while simultaneously minimizing false alarms, we report support-weighted precision, recall, and F1 across each outcome. We test the \fancyModelName{} model against the best baseline from Section \ref{behavior_prediction_results}, the ARIMA model, both working in the \paperNameActsafe{} framework. The results of our evaluation are presented in Table \ref{violation_results_table}. We see that the \fancyModelName{} model outperforms the ARIMA baseline model in each of the two MTC type tasks, and that while the \paperNameActsafe{} framework successfully predicts consistency constraint violations, the definitive dependency constraint violations are harder to predict. We hypothesize that this is due to the increased complexity in predicting both the \textit{medication intake} and \textit{eating} RHBs, as opposed to just the \textit{medication intake} RHB timestamp prediction that is solely relied upon in the consistency constraint violation prediction task.

It is important to remember the implications of correct and incorrect MTC violation predictions in an active model setting. If the \paperNameActsafe{} framework \textit{incorrectly} predicts that a violation will occur then the patient simply receives an extra reminder about their medication intake. Conversely, if the \paperNameActsafe{} framework fails to predict an actual future violation then the patient continues as the regularly would without \paperNameActsafe{}. The benefit of the good results seen in this section is that \paperNameActsafe{} is able to minimize both of these errors simultaneously.
\section{Related Work}

\subsection{Smart Systems for Treatment Adherence}

 Zhang et al. focus on medication adherence among patients who suffer from Parkinson's disease, which often causes impaired cognition or memory loss \cite{zhang2019pdmove}. Their approach, PDMove, is a smartphone-based sensing system that passively recognizes when a patient's gait has changed, which indicates that medication treatment has not been followed. PDMove passively collects data and detects medicine intake using a multi-view convolutional neural network. 
Klakegg et al. propose a near-infrared spectroscopy scanner (NIRS) for assisted medication management in elderly care \cite{klakegg2018assisted}. Their NIRS system is able to accurately identify pharmaceuticals before they are administered to patients, assisting care providers and patients in administering correct medications on schedule. 



Emi et al. formulate the problem of general purpose activity and micro-activity modeling in order to notify users about missing process steps and potentially unsafe situations \cite{emi2017quactive}. They develop QuActive, to identify partially completed activities or missing steps in overall activity processes by using a temporal probabilistic context free grammar to define activity processes. 
Another relevant system is MedRem, a medication tracking and reminder system on wearable wrist devices \cite{mondol2016medrem}. It utilizes a medication intake user confirmation session and a reminder session to help patients achieve higher levels of medication adherence using wearable devices. 

Although these above-mentioned solutions can help the users to take their medication in time or track the medication intake behavior, none of them focus on the aspect of violations of medical temporal constraints (MTCs). \paperNameActsafe{} addresses this knowledge gap. Our proposed solution can be integrated to these existing medication adherence solutions to increase the medication adherence of patients. 



\subsection{Predictive Modeling of Regular Health Behaviors (RHBs)}
Predictive modeling of human behavior is a burgeoning field, with an increased focus on sensor-based activity recognition models \cite{hao2018activity, ghods2019activity2vec, minor2015data}.
There are three main formulations of the sensor-based activity prediction task. First, it can be formulated as a next-activity prediction task in which prior sensor readings and recognized activities are used to predict which activity will occur next out of a set of activities  \cite{tax2018human, alaghbari2022activities, kim2017activity, krishna2018lstm}, all of whom utilize deep learning frameworks similar to the \fancyModelName{} model. Most similar to \paperNameActsafe{} among these is the work of Krishna et al. whose two distinct LSTM-based models predict both the next activity and the duration of that activity \cite{krishna2018lstm}. Their model also utilizes a high-level labeled activity approach of modeling prior sequences of activities and their duration. However, it differs from the \fancyModelName{} model in that \fancyModelName{} predicts the timestamp of the future activity. 
Second, the sensor-based activity prediction task can be formulated as a sensor prediction task, as shown in \cite{nazerfard2015crafft}, where the regression task is solved with Bayesian networks. The final formulation of the sensor-based activity prediction task is most similar to the \fancyModelName{} model formulation, the task of predicting the timestamp of each activity in a set, as explored by \cite{minor2017forecasting, yang2020multi}. All prior activity prediction tasks outside of Krishna et al., however, utilize prior sensor readings when predicting future activities. The \paperNameActsafe{} mobile app is adaptable to any device with an existing human activity recognition (HAR) framework, making it a flexible solution for predicting safety-critical RHBs. However, this flexibility also makes \paperNameActsafe{} to rely heavily on the ground truths of relevant behaviors or RHBs which are often harder to collect. 


Most similar to the \fancyModelName{} model is the work of Preum et al. who focus on predicting human activities in different online domains such as Twitter and search logs, as well as single-resident and multi-resident smart home environments \cite{preum2015maper}. They present a personalized activity prediction model that is adaptive to multi-scale behaviors, i.e., it models both short-span and long-span behaviors. Additionally, the MAPer model is like the \paperNameActsafe{} framework in that it utilizes high-level labeled activities, and they introduce the basis vectorization approach for representing prior behavior patterns used in the \fancyModelName{} model. However, their focus is on predicting different online and offline temporal behaviors instead of behaviors relevant to medication adherence.


Existing state-of-the-art activity forecasting models have a different setting than \paperNameActsafe{}. These models focus on sensor-based prediction of future activities, whereas in \paperNameActsafe{} we choose to work on a higher level of abstraction by utilizing only logs of labeled activities/RHBs. We choose this focus as it is a more flexible solution, capable of being utilized in any setting with a usable human activity recognition system. These settings include smart home environments, smartphone sensor readings, smart wearable devices, or some combination of these devices as is the case in the study which produced our RHB dataset, described in Section \ref{RHB_dataset}.

\section{Discussion, Limitations, and Future Work}

\paperNameActsafe{} is a proof-of-concept solution to demonstrate the feasibility of a computational pipeline for predicting violations of MTCs to increase treatment adherence. One of the \textbf{major limitations of this work} is the lack real deployment. In the future, we plan to conduct a user study to evaluate the effectiveness of \paperNameActsafe{} to increase medication adherence through predicting violations of MTCs. Other limitations and potential future work corresponding to different components of \paperNameActsafe{} are discussed below.

\subsection{MTC Extraction}
We focus on extracting MTCs from drug usage guidelines in this paper. Similar techniques for modeling and extracting medical temporal constraints could be useful for other medical regimens with temporal constraints, e.g., clinical guidelines for post-surgery recovery, rehabilitation, and physical therapy. While the substring matching algorithm is effective in mapping to MTCs using the CFG taxonomy in our drug usage guideline (DUG) dataset, some texts imply MTCs that are impossible to extract via this method. Additionally, both the algorithm and the CFG were developed based on our DUG dataset. The generalizability of the CFG and substring matching algorithm to other patient education materials is subject to future investigation. This model is also explicitly designed only for English language. Each extracted MTC should be reviewed and approved by both the patient and their health-care provider.


\subsection{Regular Health Behavior Prediction}

While the \fancyModelName{} model can predict \textit{regular} health behaviors (RHBs) with low RMSE, extremely irregular health behaviors are edge cases not handled by the model. For example, we excluded some patients' RHB logs in cases of significantly sparse or discontinued label spaces. Some rare behaviors can also be health-adjacent. For example, say a patient takes a medication which makes them dizzy, so they impose the MTC \textit{not 2 hours before driving}. They live within walking distance of most of their daily activities, however, so they only drive occasionally on weekends or for vacation. The problem of detecting this behavior is a safety critical one in the context of medication adherence. However, the \fancyModelName{} model may have difficulty predicting this pattern of behavior with limited data, if such a behavior is able to be labeled at all by the underlying activity recognition system. Additionally, there may be RHBs that are overlooked by the underlying activity recognition system. Future work in health behavior prediction will look to address both the rare and unrecognized behavior issues.

\subsection{MTC Violation}
In Section \ref{predicting_violation_results_section} we show that \paperNameActsafe{} is able to predict real MTC violations of real chronic disease patients in a real-world, uncontrolled environment. In this MTC violation prediction study we predict violations of the two MTCs associated with ACE inhibitors, however we are not able to test the \paperNameActsafe{} framework in predicting other types of MTCs since each patient in the RHB dataset is prescribed the same medication type. In the future, we plan evaluate violation prediction across other MTC types. We also note that while we have built a prototype \paperNameActsafe{} mobile application, we do not evaluate the application itself in this work. Future work will be required to evaluate the effectiveness of the \paperNameActsafe{} app to improve medication adherence. 

Although we evaluate \paperNameActsafe{} for ACE inhibitors in the context of a cardiovascular disease, our solution is generalizable to other prescription and over-the-counter medications as well. For instance, oral contraceptives commonly have the \textit{consistency MTC} (type 6) \textit{same time each day} and become ineffective when this MTC is violated, potentially with severe consequences.

While in our \paperNameActsafe{} evaluation each RHB patient was assumed to be taking only one medication, it should be noted that the \paperNameActsafe{} framework is also able to predict MTCs for multiple medications by predicting each medication event independently. This makes the assumption that each \textit{medication intake} RHB will be labeled separately by the underlying HAR system, whether it be through separate Bluetooth medication storage devices as in our RHB dataset or otherwise, and that there will be separate trained models for each of the multiple \textit{medication intake} behaviors. The process of MTC violation prediction also requires the manual step of normalizing behavior labels mentioned in the DUG data and captured by the activity recognition system. For example, a behavior or activity label for a given HAR system may label the \textit{sleep} RHB as \textit{bedroom activity}. Manual matching of such labels is required by \paperNameActsafe{} to accurately predict MTC violation.

Our formulation of \paperNameActsafe{} relies on the implicit assumption that temporal dependencies between RHBs will remain consistent over time.  Violations of this assumption would occur in situations when patients change lifestyles, either temporarily (e.g., holiday, vacation) or permanently (e.g., new job, relocation). Since \paperNameActsafe{} tracks patient behaviors, it would be aware of these schedule anomalies. One potential solution in these instances would be to prompt the patient to confirm (i) whether a schedule change has occurred, and (ii) whether it will be temporary or permanent. If temporary, \paperNameActsafe{} could utilize a simple reminder system for the duration of the temporary schedule uncertainty. If permanent, the model could be retrained in an active setting. We plan to explore these ideas for in the future.

\section{Conclusion}
In this paper we present \paperNameActsafe{}, a novel architecture for modeling, extracting and predicting violations of medical temporal constraints (MTCs) associated with regular health behaviors (RHBs) in the context of medication adherence in chronic disease patients. \paperNameActsafe{} is a proof-of-concept solution to demonstrate the feasibility of a computational pipeline for predicting violations of MTCs. Techniques based on a new context free grammar for modeling MTCs allow \paperNameActsafe{} to extract MTCs from drug usage guidelines with a weighted F1 of 0.66. Empirical evaluation shows that the HEalth Related BEhavior pRedicTion (\fancyModelName{}) model used in \paperNameActsafe{} is able to predict future regular health behaviors with low RMSE, outperforming all baselines by an average RMSE reduction of 51\%. A study based on a real-world, uncontrolled RHB dataset with $N=28$ chronic disease patients shows that \paperNameActsafe{} correctly predicts MTC violations with an average F1 score of 0.86. We believe that \paperNameActsafe{} can be a useful system with the potential to improve medication adherence and health safety.

\bibliographystyle{ACM-Reference-Format}
\bibliography{UbiComp_2022}


\begin{thebibliography}{47}


\ifx \showCODEN    \undefined \def \showCODEN     #1{\unskip}     \fi
\ifx \showDOI      \undefined \def \showDOI       #1{#1}\fi
\ifx \showISBNx    \undefined \def \showISBNx     #1{\unskip}     \fi
\ifx \showISBNxiii \undefined \def \showISBNxiii  #1{\unskip}     \fi
\ifx \showISSN     \undefined \def \showISSN      #1{\unskip}     \fi
\ifx \showLCCN     \undefined \def \showLCCN      #1{\unskip}     \fi
\ifx \shownote     \undefined \def \shownote      #1{#1}          \fi
\ifx \showarticletitle \undefined \def \showarticletitle #1{#1}   \fi
\ifx \showURL      \undefined \def \showURL       {\relax}        \fi
\providecommand\bibfield[2]{#2}
\providecommand\bibinfo[2]{#2}
\providecommand\natexlab[1]{#1}
\providecommand\showeprint[2][]{arXiv:#2}

\bibitem[may({[n.\,d.]})]%
        {mayoCACE}
 \bibinfo{year}{[n.\,d.]}\natexlab{}.
\newblock \bibinfo{title}{Angiotensin-converting enzyme (ACE) inhibitors}.
\newblock
  \bibinfo{howpublished}{\url{https://www.mayoclinic.org/diseases-conditions/high-blood-pressure/in-depth/ace-inhibitors/art-20047480
  }}.
\newblock
\newblock
\shownote{Accessed: 2022-11-10}.


\bibitem[Med({[n.\,d.]})]%
        {Medscape}
 \bibinfo{year}{[n.\,d.]}\natexlab{}.
\newblock \bibinfo{title}{Medscape: Search Drugs, OTCs \& Herbals}.
\newblock \bibinfo{howpublished}{\url{http://reference.medscape.com/drugs}}.
\newblock
\newblock
\shownote{Accessed: 2017-04-15}.


\bibitem[nio({[n.\,d.]})]%
        {nioa}
 \bibinfo{year}{[n.\,d.]}\natexlab{}.
\newblock \bibinfo{title}{Taking Medicines Safely as You Age}.
\newblock
\newblock
\urldef\tempurl%
\url{https://www.nia.nih.gov/health/taking-medicines-safely-you-age#:~:text=Taking\%20medicines\%20on\%20an\%20empty,two\%20hours\%20after\%20you\%20eat.}
\showURL{%
\tempurl}


\bibitem[CDC({[n.\,d.]})]%
        {CDCPrescription}
 \bibinfo{year}{[n.\,d.]}\natexlab{}.
\newblock \bibinfo{title}{Therapeutic Drug Use: Natiotnal Center for Health
  Statistics}.
\newblock
  \bibinfo{howpublished}{\url{https://www.cdc.gov/nchs/fastats/drug-use-therapeutic.htm}}.
\newblock
\newblock
\shownote{Accessed: 2022-04-05}.


\bibitem[mts({[n.\,d.]})]%
        {mtsamples}
 \bibinfo{year}{[n.\,d.]}\natexlab{}.
\newblock \bibinfo{title}{Transcribed Medical Transcription Sample Reports and
  Examples}.
\newblock \bibinfo{howpublished}{\url{http://www.mtsamples.com/}}.
\newblock
\newblock
\shownote{Accessed: 2017-03-15}.


\bibitem[CDC(2022)]%
        {CDCPrescription2}
 \bibinfo{year}{2022}\natexlab{}.
\newblock \bibinfo{title}{CDC Grand Rounds: Improving Medication Adherence for
  Chronic Disease Management — Innovations and Opportunities: Morbidity and
  Mortality Weekly Report (MMWR)}.
\newblock
  \bibinfo{howpublished}{\url{https://www.cdc.gov/mmwr/volumes/66/wr/mm6645a2.htm}}.
\newblock
\newblock
\shownote{Accessed: 2022-04-05}.


\bibitem[Alaghbari et~al\mbox{.}(2022)]%
        {alaghbari2022activities}
\bibfield{author}{\bibinfo{person}{Khaled~A Alaghbari},
  \bibinfo{person}{Mohamad Hanif~Md Saad}, \bibinfo{person}{Aini Hussain},
  {and} \bibinfo{person}{Muhammad~Raisul Alam}.}
  \bibinfo{year}{2022}\natexlab{}.
\newblock \showarticletitle{Activities Recognition, Anomaly Detection and Next
  Activity Prediction Based on Neural Networks in Smart Homes}.
\newblock \bibinfo{journal}{\emph{IEEE Access}}  \bibinfo{volume}{10}
  (\bibinfo{year}{2022}), \bibinfo{pages}{28219--28232}.
\newblock


\bibitem[Burnier et~al\mbox{.}(2020)]%
        {burnier2020hypertension}
\bibfield{author}{\bibinfo{person}{Michel Burnier}, \bibinfo{person}{Erietta
  Polychronopoulou}, {and} \bibinfo{person}{Gregoire Wuerzner}.}
  \bibinfo{year}{2020}\natexlab{}.
\newblock \showarticletitle{Hypertension and drug adherence in the elderly}.
\newblock \bibinfo{journal}{\emph{Frontiers in cardiovascular medicine}}
  \bibinfo{volume}{7} (\bibinfo{year}{2020}), \bibinfo{pages}{49}.
\newblock


\bibitem[Casagrande(2019)]%
        {casagrande2019sensor}
\bibfield{author}{\bibinfo{person}{Fl{\'a}via~Dias Casagrande}.}
  \bibinfo{year}{2019}\natexlab{}.
\newblock \showarticletitle{PhD Thesis: Sensor event and activity prediction
  using binary sensors in real homes with older adults}.
\newblock  (\bibinfo{year}{2019}).
\newblock


\bibitem[Chisholm-Burns and Spivey(2012)]%
        {chisholm2012cost}
\bibfield{author}{\bibinfo{person}{Marie~A Chisholm-Burns} {and}
  \bibinfo{person}{Christina~A Spivey}.} \bibinfo{year}{2012}\natexlab{}.
\newblock \showarticletitle{The'cost'of medication nonadherence: consequences
  we cannot afford to accept}.
\newblock \bibinfo{journal}{\emph{Journal of the American Pharmacists
  Association}} \bibinfo{volume}{52}, \bibinfo{number}{6}
  (\bibinfo{year}{2012}), \bibinfo{pages}{823--826}.
\newblock


\bibitem[Cook and Schmitter-Edgecombe(2009)]%
        {cook2009assessing}
\bibfield{author}{\bibinfo{person}{Diane~J Cook} {and} \bibinfo{person}{Maureen
  Schmitter-Edgecombe}.} \bibinfo{year}{2009}\natexlab{}.
\newblock \showarticletitle{Assessing the quality of activities in a smart
  environment}.
\newblock \bibinfo{journal}{\emph{Methods of information in medicine}}
  \bibinfo{volume}{48}, \bibinfo{number}{5} (\bibinfo{year}{2009}),
  \bibinfo{pages}{480}.
\newblock


\bibitem[Dahmen et~al\mbox{.}(2018)]%
        {dahmen2018smart}
\bibfield{author}{\bibinfo{person}{Jessamyn Dahmen}, \bibinfo{person}{Bryan
  Minor}, \bibinfo{person}{Diane Cook}, \bibinfo{person}{Thao Vo}, {and}
  \bibinfo{person}{Maureen Schmitter-Edgecombe}.}
  \bibinfo{year}{2018}\natexlab{}.
\newblock \showarticletitle{Smart Home-driven Digital Memory Notebook Support
  of Activity Self-Management for Older Adults}.
\newblock \bibinfo{journal}{\emph{Gerontechnology}}  \bibinfo{volume}{17}
  (\bibinfo{year}{2018}), \bibinfo{pages}{113--125}.
\newblock


\bibitem[DiMatteo(2004)]%
        {dimatteo2004variations}
\bibfield{author}{\bibinfo{person}{M~Robin DiMatteo}.}
  \bibinfo{year}{2004}\natexlab{}.
\newblock \showarticletitle{Variations in patients' adherence to medical
  recommendations: a quantitative review of 50 years of research}.
\newblock \bibinfo{journal}{\emph{Medical care}} (\bibinfo{year}{2004}),
  \bibinfo{pages}{200--209}.
\newblock


\bibitem[Dolce et~al\mbox{.}(1991)]%
        {dolce1991medication}
\bibfield{author}{\bibinfo{person}{Jeffrey~J Dolce}, \bibinfo{person}{Carol
  Crisp}, \bibinfo{person}{Bryn Manzella}, \bibinfo{person}{James~M Richards},
  \bibinfo{person}{J~Michael Hardin}, {and} \bibinfo{person}{William~C
  Bailey}.} \bibinfo{year}{1991}\natexlab{}.
\newblock \showarticletitle{Medication adherence patterns in chronic
  obstructive pulmonary disease}.
\newblock \bibinfo{journal}{\emph{Chest}} \bibinfo{volume}{99},
  \bibinfo{number}{4} (\bibinfo{year}{1991}), \bibinfo{pages}{837--841}.
\newblock


\bibitem[Donnan et~al\mbox{.}(2002)]%
        {donnan2002adherence}
\bibfield{author}{\bibinfo{person}{Peter~T Donnan}, \bibinfo{person}{Thomas~M
  MacDonald}, {and} \bibinfo{person}{Andrew~D Morris}.}
  \bibinfo{year}{2002}\natexlab{}.
\newblock \showarticletitle{Adherence to prescribed oral hypoglycaemic
  medication in a population of patients with Type 2 diabetes: a retrospective
  cohort study}.
\newblock \bibinfo{journal}{\emph{Diabetic Medicine}} \bibinfo{volume}{19},
  \bibinfo{number}{4} (\bibinfo{year}{2002}), \bibinfo{pages}{279--284}.
\newblock


\bibitem[Eldesokey et~al\mbox{.}(2018)]%
        {eldesokey2018propagating}
\bibfield{author}{\bibinfo{person}{Abdelrahman Eldesokey},
  \bibinfo{person}{Michael Felsberg}, {and} \bibinfo{person}{Fahad~Shahbaz
  Khan}.} \bibinfo{year}{2018}\natexlab{}.
\newblock \showarticletitle{Propagating confidences through cnns for sparse
  data regression}.
\newblock \bibinfo{journal}{\emph{arXiv preprint arXiv:1805.11913}}
  (\bibinfo{year}{2018}).
\newblock


\bibitem[Emi et~al\mbox{.}(2017)]%
        {emi2017quactive}
\bibfield{author}{\bibinfo{person}{Ifat~Afrin Emi},
  \bibinfo{person}{Md~Abu~Sayeed Mondol}, {and} \bibinfo{person}{John~A
  Stankovic}.} \bibinfo{year}{2017}\natexlab{}.
\newblock \showarticletitle{QuActive: a quality of activities monitoring and
  notification system}. In \bibinfo{booktitle}{\emph{Proceedings of the 8th
  International Conference on Cyber-Physical Systems}}. ACM,
  \bibinfo{pages}{281--291}.
\newblock


\bibitem[Ferguson et~al\mbox{.}(2017)]%
        {ferguson2017barriers}
\bibfield{author}{\bibinfo{person}{Caleb Ferguson}, \bibinfo{person}{Sally~C
  Inglis}, \bibinfo{person}{Phillip~J Newton}, \bibinfo{person}{Sandy
  Middleton}, \bibinfo{person}{Peter~S Macdonald}, {and}
  \bibinfo{person}{Patricia~M Davidson}.} \bibinfo{year}{2017}\natexlab{}.
\newblock \showarticletitle{Barriers and enablers to adherence to
  anticoagulation in heart failure with atrial fibrillation: patient and
  provider perspectives}.
\newblock \bibinfo{journal}{\emph{Journal of clinical nursing}}
  \bibinfo{volume}{26}, \bibinfo{number}{23-24} (\bibinfo{year}{2017}),
  \bibinfo{pages}{4325--4334}.
\newblock


\bibitem[Frazee et~al\mbox{.}(2012)]%
        {expressScripts}
\bibfield{author}{\bibinfo{person}{Sharon Frazee}, \bibinfo{person}{Steven
  Miller}, \bibinfo{person}{Robert Nease}, {and} \bibinfo{person}{Glen
  Stettin}.} \bibinfo{year}{2012}\natexlab{}.
\newblock \bibinfo{booktitle}{\emph{2011 Drug Trend Report}}.
\newblock \bibinfo{type}{{T}echnical {R}eport}. \bibinfo{institution}{The
  EXPRESS SCRIPTS RESEARCH \& NEW SOLUTIONS LAB}.
\newblock


\bibitem[Ghods and Cook(2019)]%
        {ghods2019activity2vec}
\bibfield{author}{\bibinfo{person}{Alireza Ghods} {and}
  \bibinfo{person}{Diane~J Cook}.} \bibinfo{year}{2019}\natexlab{}.
\newblock \showarticletitle{Activity2Vec: Learning ADL Embeddings from Sensor
  Data with a Sequence-to-Sequence Model}.
\newblock \bibinfo{journal}{\emph{arXiv preprint arXiv:1907.05597}}
  (\bibinfo{year}{2019}).
\newblock


\bibitem[Ha et~al\mbox{.}(2014)]%
        {ha2014towards}
\bibfield{author}{\bibinfo{person}{Kiryong Ha}, \bibinfo{person}{Zhuo Chen},
  \bibinfo{person}{Wenlu Hu}, \bibinfo{person}{Wolfgang Richter},
  \bibinfo{person}{Padmanabhan Pillai}, {and} \bibinfo{person}{Mahadev
  Satyanarayanan}.} \bibinfo{year}{2014}\natexlab{}.
\newblock \showarticletitle{Towards wearable cognitive assistance}. In
  \bibinfo{booktitle}{\emph{Proceedings of the 12th annual international
  conference on Mobile systems, applications, and services}}. ACM,
  \bibinfo{pages}{68--81}.
\newblock


\bibitem[Hao et~al\mbox{.}(2018)]%
        {hao2018activity}
\bibfield{author}{\bibinfo{person}{Jianguo Hao}, \bibinfo{person}{Abdenour
  Bouzouane}, \bibinfo{person}{Bruno Bouchard}, {and}
  \bibinfo{person}{S{\'e}bastien Gaboury}.} \bibinfo{year}{2018}\natexlab{}.
\newblock \showarticletitle{Activity inference engine for real-time cognitive
  assistance in smart environments}.
\newblock \bibinfo{journal}{\emph{Journal of Ambient Intelligence and Humanized
  Computing}} \bibinfo{volume}{9}, \bibinfo{number}{3} (\bibinfo{year}{2018}),
  \bibinfo{pages}{679--698}.
\newblock


\bibitem[Hinkin et~al\mbox{.}(2002)]%
        {hinkin2002medication}
\bibfield{author}{\bibinfo{person}{CH Hinkin}, \bibinfo{person}{SA Castellon},
  \bibinfo{person}{RS Durvasula}, \bibinfo{person}{DJ Hardy},
  \bibinfo{person}{MN Lam}, \bibinfo{person}{KI Mason}, \bibinfo{person}{D
  Thrasher}, \bibinfo{person}{MB Goetz}, {and} \bibinfo{person}{M Stefaniak}.}
  \bibinfo{year}{2002}\natexlab{}.
\newblock \showarticletitle{Medication adherence among HIV+ adults: effects of
  cognitive dysfunction and regimen complexity}.
\newblock \bibinfo{journal}{\emph{Neurology}} \bibinfo{volume}{59},
  \bibinfo{number}{12} (\bibinfo{year}{2002}), \bibinfo{pages}{1944--1950}.
\newblock


\bibitem[Ingersoll and Cohen(2008)]%
        {ingersoll2008impact}
\bibfield{author}{\bibinfo{person}{Karen~S Ingersoll} {and}
  \bibinfo{person}{Jessye Cohen}.} \bibinfo{year}{2008}\natexlab{}.
\newblock \showarticletitle{The impact of medication regimen factors on
  adherence to chronic treatment: a review of literature}.
\newblock \bibinfo{journal}{\emph{Journal of behavioral medicine}}
  \bibinfo{volume}{31}, \bibinfo{number}{3} (\bibinfo{year}{2008}),
  \bibinfo{pages}{213--224}.
\newblock


\bibitem[Kendall and Ord(1990)]%
        {arima}
\bibfield{author}{\bibinfo{person}{Maurice~George Kendall} {and}
  \bibinfo{person}{John~Keith Ord}.} \bibinfo{year}{1990}\natexlab{}.
\newblock \bibinfo{booktitle}{\emph{Time-series}}. Vol.~\bibinfo{volume}{296}.
\newblock \bibinfo{publisher}{Edward Arnold London}.
\newblock


\bibitem[Kim et~al\mbox{.}(2017)]%
        {kim2017activity}
\bibfield{author}{\bibinfo{person}{Younggi Kim}, \bibinfo{person}{Jihoon An},
  \bibinfo{person}{Minseok Lee}, {and} \bibinfo{person}{Younghee Lee}.}
  \bibinfo{year}{2017}\natexlab{}.
\newblock \showarticletitle{An activity-embedding approach for next-activity
  prediction in a multi-user smart space}. In \bibinfo{booktitle}{\emph{2017
  IEEE International Conference on Smart Computing (SMARTCOMP)}}. IEEE,
  \bibinfo{pages}{1--6}.
\newblock


\bibitem[Klakegg et~al\mbox{.}(2018)]%
        {klakegg2018assisted}
\bibfield{author}{\bibinfo{person}{Simon Klakegg}, \bibinfo{person}{Jorge
  Goncalves}, \bibinfo{person}{Chu Luo}, \bibinfo{person}{Aku Visuri},
  \bibinfo{person}{Alexey Popov}, \bibinfo{person}{Niels van Berkel},
  \bibinfo{person}{Zhanna Sarsenbayeva}, \bibinfo{person}{Vassilis Kostakos},
  \bibinfo{person}{Simo Hosio}, \bibinfo{person}{Scott Savage},
  {et~al\mbox{.}}} \bibinfo{year}{2018}\natexlab{}.
\newblock \showarticletitle{Assisted medication management in elderly care
  using miniaturised near-infrared spectroscopy}.
\newblock \bibinfo{journal}{\emph{Proceedings of the ACM on Interactive,
  Mobile, Wearable and Ubiquitous Technologies}} \bibinfo{volume}{2},
  \bibinfo{number}{2} (\bibinfo{year}{2018}), \bibinfo{pages}{1--24}.
\newblock


\bibitem[Krishna et~al\mbox{.}(2018)]%
        {krishna2018lstm}
\bibfield{author}{\bibinfo{person}{Kundan Krishna}, \bibinfo{person}{Deepali
  Jain}, \bibinfo{person}{Sanket~V Mehta}, {and} \bibinfo{person}{Sunav
  Choudhary}.} \bibinfo{year}{2018}\natexlab{}.
\newblock \showarticletitle{An lstm based system for prediction of human
  activities with durations}.
\newblock \bibinfo{journal}{\emph{Proceedings of the ACM on Interactive,
  Mobile, Wearable and Ubiquitous Technologies}} \bibinfo{volume}{1},
  \bibinfo{number}{4} (\bibinfo{year}{2018}), \bibinfo{pages}{1--31}.
\newblock


\bibitem[Leven et~al\mbox{.}(2017)]%
        {leven2017medication}
\bibfield{author}{\bibinfo{person}{Emily~A Leven}, \bibinfo{person}{Rachel
  Annunziato}, \bibinfo{person}{Jacqueline Helcer}, \bibinfo{person}{Sarah~R
  Lieber}, \bibinfo{person}{Christopher~S Knight}, \bibinfo{person}{Catherine
  Wlodarkiewicz}, \bibinfo{person}{Rainier~P Soriano},
  \bibinfo{person}{Sander~S Florman}, \bibinfo{person}{Thomas~D Schiano}, {and}
  \bibinfo{person}{Eyal Shemesh}.} \bibinfo{year}{2017}\natexlab{}.
\newblock \showarticletitle{Medication adherence and rejection rates in older
  vs younger adult liver transplant recipients}.
\newblock \bibinfo{journal}{\emph{Clinical transplantation}}
  \bibinfo{volume}{31}, \bibinfo{number}{6} (\bibinfo{year}{2017}),
  \bibinfo{pages}{e12981}.
\newblock


\bibitem[Mayer and Stone(2001)]%
        {mayer2001strategies}
\bibfield{author}{\bibinfo{person}{Kenneth~H Mayer} {and}
  \bibinfo{person}{Valerie~E Stone}.} \bibinfo{year}{2001}\natexlab{}.
\newblock \showarticletitle{Strategies for optimizing adherence to highly
  active antiretroviral therapy: lessons from research and clinical practice}.
\newblock \bibinfo{journal}{\emph{Clinical Infectious Diseases}}
  \bibinfo{volume}{33}, \bibinfo{number}{6} (\bibinfo{year}{2001}),
  \bibinfo{pages}{865--872}.
\newblock


\bibitem[Minor and Cook(2017)]%
        {minor2017forecasting}
\bibfield{author}{\bibinfo{person}{Bryan Minor} {and} \bibinfo{person}{Diane~J
  Cook}.} \bibinfo{year}{2017}\natexlab{}.
\newblock \showarticletitle{Forecasting occurrences of activities}.
\newblock \bibinfo{journal}{\emph{Pervasive and mobile computing}}
  \bibinfo{volume}{38} (\bibinfo{year}{2017}), \bibinfo{pages}{77--91}.
\newblock


\bibitem[Minor et~al\mbox{.}(2015)]%
        {minor2015data}
\bibfield{author}{\bibinfo{person}{Bryan Minor}, \bibinfo{person}{Janardhan~Rao
  Doppa}, {and} \bibinfo{person}{Diane~J Cook}.}
  \bibinfo{year}{2015}\natexlab{}.
\newblock \showarticletitle{Data-driven activity prediction: Algorithms,
  evaluation methodology, and applications}. In
  \bibinfo{booktitle}{\emph{Proceedings of the 21th ACM SIGKDD International
  Conference on Knowledge Discovery and Data Mining}}. ACM,
  \bibinfo{pages}{805--814}.
\newblock


\bibitem[Mondol et~al\mbox{.}(2016)]%
        {mondol2016medrem}
\bibfield{author}{\bibinfo{person}{Abu~Sayeed Mondol},
  \bibinfo{person}{Ifat~Afrin Emi}, {and} \bibinfo{person}{John~A Stankovic}.}
  \bibinfo{year}{2016}\natexlab{}.
\newblock \showarticletitle{MedRem: An interactive medication reminder and
  tracking system on wrist devices}. In \bibinfo{booktitle}{\emph{2016 IEEE
  Wireless Health (WH)}}. IEEE, \bibinfo{pages}{1--8}.
\newblock


\bibitem[Nasution(2006)]%
        {nasution2006use}
\bibfield{author}{\bibinfo{person}{Sally~Aman Nasution}.}
  \bibinfo{year}{2006}\natexlab{}.
\newblock \showarticletitle{The use of ACE inhibitor in cardiovascular
  disease.}
\newblock \bibinfo{journal}{\emph{Acta Med Indones}} (\bibinfo{year}{2006}).
\newblock


\bibitem[Nazerfard and Cook(2015)]%
        {nazerfard2015crafft}
\bibfield{author}{\bibinfo{person}{Ehsan Nazerfard} {and}
  \bibinfo{person}{Diane~J Cook}.} \bibinfo{year}{2015}\natexlab{}.
\newblock \showarticletitle{CRAFFT: an activity prediction model based on
  Bayesian networks}.
\newblock \bibinfo{journal}{\emph{Journal of ambient intelligence and humanized
  computing}} \bibinfo{volume}{6}, \bibinfo{number}{2} (\bibinfo{year}{2015}),
  \bibinfo{pages}{193--205}.
\newblock


\bibitem[Osterberg and Blaschke(2005)]%
        {osterberg2005adherence}
\bibfield{author}{\bibinfo{person}{Lars Osterberg} {and}
  \bibinfo{person}{Terrence Blaschke}.} \bibinfo{year}{2005}\natexlab{}.
\newblock \showarticletitle{Adherence to medication}.
\newblock \bibinfo{journal}{\emph{New England journal of medicine}}
  \bibinfo{volume}{353}, \bibinfo{number}{5} (\bibinfo{year}{2005}),
  \bibinfo{pages}{487--497}.
\newblock


\bibitem[Paes et~al\mbox{.}(1997)]%
        {paes1997impact}
\bibfield{author}{\bibinfo{person}{Arsenio~HP Paes}, \bibinfo{person}{Albert
  Bakker}, {and} \bibinfo{person}{Carmen~J Soe-Agnie}.}
  \bibinfo{year}{1997}\natexlab{}.
\newblock \showarticletitle{Impact of dosage frequency on patient compliance}.
\newblock \bibinfo{journal}{\emph{Diabetes care}} \bibinfo{volume}{20},
  \bibinfo{number}{10} (\bibinfo{year}{1997}), \bibinfo{pages}{1512--1517}.
\newblock


\bibitem[Pham et~al\mbox{.}(2011)]%
        {pham2011national}
\bibfield{author}{\bibinfo{person}{Julius~Cuong Pham}, \bibinfo{person}{Julie~L
  Story}, \bibinfo{person}{Rodney~W Hicks}, \bibinfo{person}{Andrew~D Shore},
  \bibinfo{person}{Laura~L Morlock}, \bibinfo{person}{Dickson~S Cheung},
  \bibinfo{person}{Gabor~D Kelen}, {and} \bibinfo{person}{Peter~J Pronovost}.}
  \bibinfo{year}{2011}\natexlab{}.
\newblock \showarticletitle{National study on the frequency, types, causes, and
  consequences of voluntarily reported emergency department medication errors}.
\newblock \bibinfo{journal}{\emph{The Journal of emergency medicine}}
  \bibinfo{volume}{40}, \bibinfo{number}{5} (\bibinfo{year}{2011}),
  \bibinfo{pages}{485--492}.
\newblock


\bibitem[Pollack et~al\mbox{.}(2002)]%
        {pollack2002pearl}
\bibfield{author}{\bibinfo{person}{Martha~E Pollack}, \bibinfo{person}{Laura
  Brown}, \bibinfo{person}{Dirk Colbry}, \bibinfo{person}{Cheryl Orosz},
  \bibinfo{person}{Bart Peintner}, \bibinfo{person}{Sailesh Ramakrishnan},
  \bibinfo{person}{Sandra Engberg}, \bibinfo{person}{Judith~T Matthews},
  \bibinfo{person}{Jacqueline Dunbar-Jacob}, \bibinfo{person}{Colleen~E
  McCarthy}, {et~al\mbox{.}}} \bibinfo{year}{2002}\natexlab{}.
\newblock \showarticletitle{Pearl: A mobile robotic assistant for the elderly}.
  In \bibinfo{booktitle}{\emph{AAAI workshop on automation as eldercare}},
  Vol.~\bibinfo{volume}{2002}. \bibinfo{pages}{85--91}.
\newblock


\bibitem[Preum et~al\mbox{.}(2018)]%
        {preum2018corpus}
\bibfield{author}{\bibinfo{person}{Sarah~Masud Preum},
  \bibinfo{person}{Md~Rizwan Parvez}, \bibinfo{person}{Kai-Wei Chang}, {and}
  \bibinfo{person}{John Stankovic}.} \bibinfo{year}{2018}\natexlab{}.
\newblock \showarticletitle{A corpus of drug usage guidelines annotated with
  type of advice}. In \bibinfo{booktitle}{\emph{Proceedings of the Eleventh
  International Conference on Language Resources and Evaluation (LREC 2018)}}.
\newblock


\bibitem[Preum et~al\mbox{.}(2015)]%
        {preum2015maper}
\bibfield{author}{\bibinfo{person}{Sarah~Masud Preum}, \bibinfo{person}{John~A
  Stankovic}, {and} \bibinfo{person}{Yanjun Qi}.}
  \bibinfo{year}{2015}\natexlab{}.
\newblock \showarticletitle{MAPer: a multi-scale adaptive personalized model
  for temporal human behavior prediction}. In
  \bibinfo{booktitle}{\emph{Proceedings of the 24th ACM International on
  Conference on Information and Knowledge Management}}. ACM,
  \bibinfo{pages}{433--442}.
\newblock


\bibitem[Rajanna et~al\mbox{.}(2014)]%
        {rajanna2014step}
\bibfield{author}{\bibinfo{person}{Vijay Rajanna}, \bibinfo{person}{Raniero
  Lara-Garduno}, \bibinfo{person}{Dev~Jyoti Behera}, \bibinfo{person}{Karthic
  Madanagopal}, \bibinfo{person}{Daniel Goldberg}, {and} \bibinfo{person}{Tracy
  Hammond}.} \bibinfo{year}{2014}\natexlab{}.
\newblock \showarticletitle{Step up life: a context aware health assistant}. In
  \bibinfo{booktitle}{\emph{Proceedings of the Third ACM SIGSPATIAL
  International Workshop on the Use of GIS in Public Health}}. ACM,
  \bibinfo{pages}{21--30}.
\newblock


\bibitem[Singh and Varshney(2019)]%
        {singh2019reminders}
\bibfield{author}{\bibinfo{person}{Neetu Singh} {and} \bibinfo{person}{Upkar
  Varshney}.} \bibinfo{year}{2019}\natexlab{}.
\newblock \showarticletitle{Reminders for medication adherence: a review and
  research agenda}.
\newblock \bibinfo{journal}{\emph{International Journal of Electronic
  Healthcare}} \bibinfo{volume}{11}, \bibinfo{number}{1}
  (\bibinfo{year}{2019}), \bibinfo{pages}{25--44}.
\newblock


\bibitem[Sonntag(2015)]%
        {sonntag2015kognit}
\bibfield{author}{\bibinfo{person}{Daniel Sonntag}.}
  \bibinfo{year}{2015}\natexlab{}.
\newblock \showarticletitle{Kognit: Intelligent cognitive enhancement
  technology by cognitive models and mixed reality for dementia patients}. In
  \bibinfo{booktitle}{\emph{2015 AAAI Fall Symposium Series}}.
\newblock


\bibitem[Tax(2018)]%
        {tax2018human}
\bibfield{author}{\bibinfo{person}{Niek Tax}.} \bibinfo{year}{2018}\natexlab{}.
\newblock \showarticletitle{Human activity prediction in smart home
  environments with LSTM neural networks}. In \bibinfo{booktitle}{\emph{2018
  14th International Conference on Intelligent Environments (IE)}}. IEEE,
  \bibinfo{pages}{40--47}.
\newblock


\bibitem[Yang et~al\mbox{.}(2020)]%
        {yang2020multi}
\bibfield{author}{\bibinfo{person}{Hong Yang}, \bibinfo{person}{Shanshan Gong},
  \bibinfo{person}{Yaqing Liu}, \bibinfo{person}{Zhengkui Lin}, {and}
  \bibinfo{person}{Yi Qu}.} \bibinfo{year}{2020}\natexlab{}.
\newblock \showarticletitle{A multi-task learning model for daily activity
  forecast in smart home}.
\newblock \bibinfo{journal}{\emph{Sensors}} \bibinfo{volume}{20},
  \bibinfo{number}{7} (\bibinfo{year}{2020}), \bibinfo{pages}{1933}.
\newblock


\bibitem[Zhang et~al\mbox{.}(2019)]%
        {zhang2019pdmove}
\bibfield{author}{\bibinfo{person}{Hanbin Zhang}, \bibinfo{person}{Chenhan Xu},
  \bibinfo{person}{Huining Li}, \bibinfo{person}{Aditya~Singh Rathore},
  \bibinfo{person}{Chen Song}, \bibinfo{person}{Zhisheng Yan},
  \bibinfo{person}{Dongmei Li}, \bibinfo{person}{Feng Lin},
  \bibinfo{person}{Kun Wang}, {and} \bibinfo{person}{Wenyao Xu}.}
  \bibinfo{year}{2019}\natexlab{}.
\newblock \showarticletitle{Pdmove: Towards passive medication adherence
  monitoring of parkinson's disease using smartphone-based gait assessment}.
\newblock \bibinfo{journal}{\emph{Proceedings of the ACM on interactive,
  mobile, wearable and ubiquitous technologies}} \bibinfo{volume}{3},
  \bibinfo{number}{3} (\bibinfo{year}{2019}), \bibinfo{pages}{1--23}.
\newblock


\end{thebibliography}

\appendix
\section{Appendix}

\subsection{Basis Vectorization Details}
\label{appendix_basis_vec}
In Algorithm \ref{basis_vectorization_algorithm} we present psuedocode for the basis vectorization algorithm described in Section \ref{ActPredSolution}. This algorithm has three parameters; an RHB log $D$, a set of recorded RHBs $B$, and the window size $x$. Algorithm \ref{basis_vectorization_algorithm} defines a function $\phi: (D,B,x) \xrightarrow{} \mathbb{R}^{M \times K}$ where $M = |B|$ and $K$ is the number of $x$-minute windows between $t_{start_1}$ and $t_{stop_{Q}}$, which defines a matrix $A \in \mathbb{R}^{M \times K}$. Each element $a_{i,j} \in A$ is either a $1$ if behavior $i$ was recorded in the $j$th window, or $0$ otherwise.

\begin{algorithm}
\caption{Basis Vectorization}
\label{basis_vectorization_algorithm}
\begin{algorithmic}
\Require RHB Log $D$, Set of RHBs $B$, Window Size $x$
\State $Q \gets |D|$
\State $K \gets \frac{(t_{stop_{Q}} - t_{start_1})}{x}$ (in minutes)
\State $M \gets |B|$
\State $BV \gets MxK$ matrix of $0$s
\State $w_{start} \gets t_{start_1}$
\For{$j$ in $0 \ldots M$}
    \For{$i$ in $0 \ldots K$}
        \If{User participated in RHB $b_j$ between time $w_{start}$ and $w_{start} + x$}
            \State $BV_{i,j} \gets 1$ 
        \EndIf
    \State $w_{start} \gets w_{start} + x$
    \EndFor
\EndFor \\
\Return $BV$
\end{algorithmic}
\end{algorithm}

\subsection{RHB Log Details and Example}
\label{appendix_rhb_log}
The RHB dataset described in Section \ref{RHB_dataset} is drawn from $n=40$ patients diagnosed with atherosclerotic cardiovascular disease (ACD) and prescribed to take a daily medication namely ACE inhibitors to manage their condition. Details of this sample of $n=40$ patients are given below.
\begin{itemize}
    \item Gender: 16 male, 24 female
    \item Age: mean 52.2 years, standard deviation 14.8 years
    \item Ethnicity:  50\% Asians, 30\% Caucasians, 10\% Hispanic, 10\% African-Americans
    \item Education: 67\% were university graduates or higher
\end{itemize}

In Table \ref{rhb_examples_table} we provide an example of an RHB log. This is a sample of 6 sequential RHBs logged by patient 141's android smartphone. It is a collection of entries $e_j = \{b_i, t_{start_j}, t_{stop_j}\}$, where each entry is associated with a specific RHB $b_i$, a start time $t_{start_j}$ and a stop time $t_{stop_j}$. Since many of the entries have short time spans, we utilize the binary basis vectorization process described in Section \ref{ActPredSolution} for reducing noise in sensor labeling.

\begin{table}
\caption{\textbf{RHB Dataset Sample}: This is a sample of 6 sequential RHBs logged by patient 141's android smartphone. Note that each entry $e$ in the dataset is associated with an RHB $b$, a start time and a stop time. Since many of the entries have short time spans, we utilize the binary basis vectorization process described in Section \ref{ActPredSolution} for reducing noise in sensor labeling.}
\label{rhb_examples_table}
\begin{tabular}{lrrr}\toprule
RHB &Start Time &Stop Time \\\midrule
work &Tue Jul 16 2019 00:21:43 &Tue Jul 16 2019 00:22:21 \\
exercise &Tue Jul 16 2019 00:20:37 &Tue Jul 16 2019 00:20:37 \\
eat &Mon Jul 15 2019 20:41:35 &Mon Jul 15 2019 20:43:14 \\
drive &Tue Jul 16 2019 00:12:53 &Tue Jul 16 2019 00:12:53 \\
work &Mon Jul 15 2019 22:38:09 &Mon Jul 15 2019 23:06:26 \\
take medicine &Mon Jul 15 2019 16:59:05 &Mon Jul 15 2019 17:00:23 \\
\bottomrule
\end{tabular}
\end{table}

\subsection{User-Level Behavior Prediction Results}
\label{appendix_tables}

In this section we provide a supplemental sample table of user-level results for the \textit{medication intake} regular health behavior (RHB) prediction task. The average results over all users in the RHB datasets are presented in Section \ref{medintake_baselines}. While the full results table is too large to be included, we provide a sample results table across 10 users, along with their user ID. As in Section \ref{medintake_baselines}, we compare the performance of \fancyModelName{} with baseline models across 3 different $x$-minute window sizes: $x \in \{15, 30, 60\}$. 

\begin{table}[h]
\caption{\textit{User-level Medication Intake RHB Prediction Results Sample:} We provide a sample results table across 10 users from the RHB datasets as described in Section \ref{RHB_dataset}. For each user (designated by the User ID column) we have three rows, each representing a different $x$-minute window size. In each row we give RMSE test results from the three baseline models; Prior-Day, ARIMA, and LSTM, as well as the \fancyModelName{} results. We also include the size of the basis vectorized training sequence (in number of $x$-minute windows), the number of training frames, and the number of testing frames for each user, $x$-minute window size pair.}
\label{user_level_results}
\begin{tabular}{lrrrrrrrrr}\toprule
ID &X-Minute &Prior-Day &ARIMA &LSTM &HERBERT &Train Seq Len &Train Frames &Test Frames \\\midrule
110 &15 &70.3302 &\textbf{57.8771} &305.9961 &68.2293 &2016 &928 &1728 \\
110 &30 &44.9081 &30.2135 &152.9980 &\textbf{29.0143} &1008 &448 &864 \\
110 &60 &17.5364 &\textbf{14.8097} &76.4990 &17.4272 &504 &224 &416 \\
112 &15 &140.4446 &\textbf{46.4438} &148.5024 &166.0748 &2016 &1344 &3424 \\
112 &30 &91.4294 &25.6409 &74.2512 &\textbf{17.4974} &1008 &672 &1696 \\
112 &60 &35.0190 &\textbf{12.4461} &37.1256 &36.8264 &504 &320 &832 \\
115 &15 &203.1004 &143.4692 &229.7867 &\textbf{131.2573} &1770 &480 &224 \\
115 &30 &134.6641 &73.4384 &114.8933 &\textbf{42.2985} &885 &224 &128 \\
115 &60 &50.7502 &36.5142 &57.4467 &\textbf{31.0895} &443 &96 &64 \\
116 &15 &97.5872 &\textbf{74.6339} &423.7276 &106.0351 &2016 &672 &704 \\
116 &30 &55.8323 &43.9542 &211.8638 &\textbf{39.3169} &1008 &320 &352 \\
116 &60 &24.3718 &\textbf{22.0133} &105.9319 &24.8992 &504 &160 &160 \\
119 &15 &\textbf{157.8849} &362.0278 &220.6434 &288.8627 &2016 &640 &640 \\
119 &30 &110.5026 &155.4700 &110.3217 &\textbf{78.7875} &1008 &320 &320 \\
119 &60 &\textbf{39.5387} &78.1531 &55.1608 &51.4075 &504 &160 &160 \\
121 &15 &140.8419 &56.2802 &342.5229 &\textbf{55.2501} &1155 &320 &160 \\
121 &30 &91.7834 &29.0433 &171.2614 &\textbf{15.4569} &578 &160 &64 \\
121 &60 &35.2371 &16.4308 &85.6307 &\textbf{4.3184} &289 &64 &32 \\
122 &15 &150.5518 &98.7327 &369.6495 &\textbf{30.2099} &2016 &1184 &2752 \\
122 &30 &100.1850 &68.1156 &184.8247 &\textbf{16.6733} &1008 &576 &1376 \\
122 &60 &37.6109 &36.9607 &92.4124 &\textbf{11.7466} &504 &288 &672 \\
123 &15 &80.3264 &46.7447 &\textbf{29.6761} &87.0045 &1481 &416 &192 \\
123 &30 &63.0530 &26.5494 &\textbf{14.8381} &31.3146 &741 &192 &96 \\
123 &60 &20.0990 &13.3471 &\textbf{7.4190} &15.4112 &371 &96 &32 \\
127 &15 &127.3313 &111.7977 &586.0778 &\textbf{31.8198} &2016 &768 &1088 \\
127 &30 &78.1049 &79.1933 &293.0389 &\textbf{20.1380} &1008 &384 &544 \\
127 &60 &31.8094 &31.5753 &146.5194 &\textbf{9.9382} &504 &192 &256 \\
128 &15 &71.8279 &31.1413 &511.3292 &\textbf{27.0651} &1751 &480 &224 \\
128 &30 &43.6295 &16.3268 &255.6646 &\textbf{9.9727} &875 &224 &96 \\
128 &60 &17.9337 &\textbf{8.8637} &127.8323 &11.2819 &438 &96 &64 \\
\bottomrule
\end{tabular}
\end{table}

In Table \ref{user_level_results} we see a sample of the user-level RMSE results from each model when predicting the \textit{medication intake} RHB. We include the size of the training sequence (in number of $x$-minute windows), the number of training frames, and the number of testing frames in each RHB dataset. We train each model on the training frames from each RHB user and present root mean squared error (RMSE) results over the test frames. Since there are $28$ RHB users and $3$ $x$-minute window sizes, there are $84$ possible configurations in these user-level results. \fancyModelName{} outperforms both baselines in $40$ of the possible $84$ configurations, whereas the ARIMA model outperforms in $30$ of the possible configurations. The Prior-Day baseline performs best in $11$ of the possible configurations, and the LSTM baseline performs best in only $3$. We see that while it may be possible or preferable for certain user schedules to perform RHB prediction using simple statistical methods or without basis vectorization, on average and in the most common case of RHB prediction, the \fancyModelName{} model is the best method for RHB prediction, making it the best option for the \paperNameActsafe{} framework.

\subsection{Schedule Regularity Details}
\label{appendix_schedule_regularity}

\begin{figure*}[htbp]
\caption{Heatmap of cosine similarity between each medication schedule in the RHB dataset. Each cell represents the similarity between user 1's medication schedule (x-axis) and user 2's medication schedule (y-axis). The higher the value, the lighter the cell color and the more similar the schedules. Note that since the $sim(x,y) = sim(y,x)$, the heatmap is symmetrical, and since $sim(x,x) = 1$, the values on the diagonal are all $1$.}
\label{regularity_heatmap_figure}
\centerline{\includegraphics[width=4in]{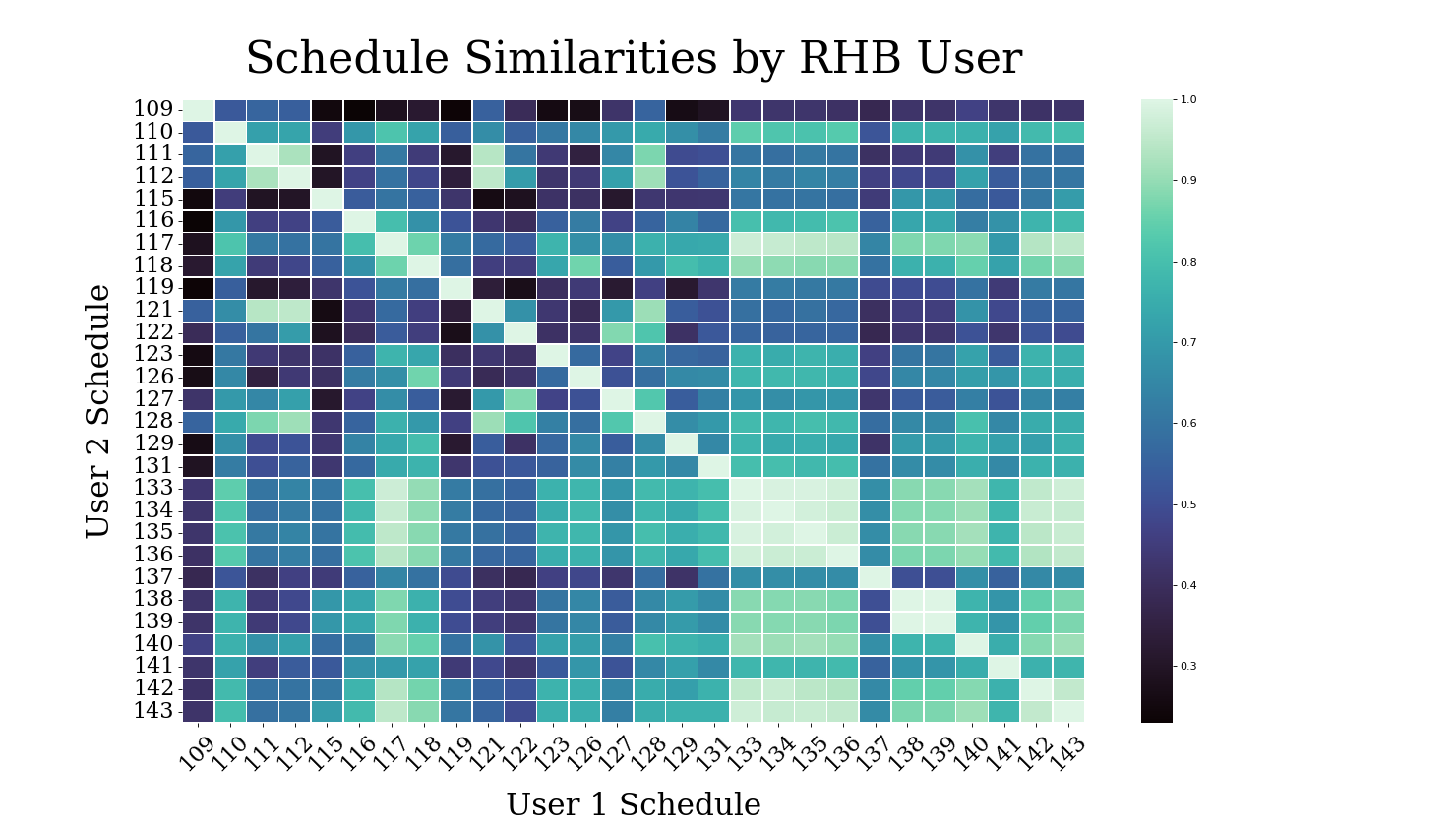}}
\end{figure*}

Running pairwise comparison between each of the users in the RHB dataset, we obtain the similarity heatmap in Figure \ref{regularity_heatmap_figure}. Since $sim(x,y) = sim(y,x)$ this heatmap is symmetrical, although we refer to users as having rows. A darker row indicates that user has a different medication schedule than other users on average. We see that to be the case specifically for user 109, who has the most unique medication schedule.

\end{document}